\documentclass{article} % For LaTeX2e
\usepackage{colm2024_conference}

\usepackage{microtype}
\usepackage{hyperref}
\usepackage{url}
\usepackage{booktabs}
\usepackage{hyperref}
\usepackage{xcolor}         % colors
\usepackage{multirow}
\usepackage{diagbox}
\usepackage{subfigure}
\usepackage{bbm}
\usepackage{xspace}
\usepackage{enumitem}
\usepackage{tikz}
\usepackage{amsmath}
\usepackage{mathrsfs}
\usepackage{dsfont}
\usepackage{amssymb}  
\usepackage{makecell}
\usepackage{algorithm}
\usepackage{algpseudocode}
\usepackage{amsthm}
\usepackage{longtable}
\usepackage{wrapfig}
\usepackage{enumitem}
\usepackage{etoc}
\usepackage{amsfonts}       % blackboard math symbols
\usepackage{nicefrac}       % compact symbols for 1/2, etc.
\usepackage{microtype}      % microtypography
\setlist{leftmargin=5mm}
\usepackage[export]{adjustbox}
\usepackage{mathtools}
\usepackage[capitalize]{cleveref}
\usepackage[textsize=tiny]{todonotes}

\theoremstyle{definition}
\newtheorem{definition}{Definition}[section]

\title{PromptRobust: Towards Evaluating the Robustness of Large Language Models on Adversarial Prompts}

\colmfinalcopy

\author{%
  Kaijie Zhu$^{1,2}$\thanks{Work done during internship at Microsoft Research Asia.}, Jindong Wang$^1$\thanks{Corresponding author: Jindong Wang <jindong.wang@microsoft.com>.}, Jiaheng Zhou$^2$, Zeek Wang$^1$, Hao Chen$^3$, Yidong Wang$^4$,\\ \textbf{Linyi Yang}$^5$, \textbf{Wei Ye}$^4$, \textbf{Neil Zhenqiang Gong}$^6$, \textbf{Yue Zhang}$^5$, \textbf{Xing Xie}$^1$ \\
  $^1$Microsoft Research ~~ $^2$Institute of Automation, CAS ~~ $^3$Carnegie Mellon University\\
  $^4$Peking University ~~ $^5$Westlake University ~~ $^6$Duke University
}

\usepackage{todonotes}
\usepackage{xcolor}

\newcommand{\method}{PromptRobust\xspace}
\newcommand{\prompt}[1]{{\ttfamily #1}\xspace}

\newcommand{\sst}{SST-2\xspace}
\newcommand{\cola}{CoLA\xspace}
\newcommand{\mnli}{MNLI\xspace}
\newcommand{\qnli}{QNLI\xspace}
\newcommand{\rte}{RTE\xspace}
\newcommand{\wnli}{WNLI\xspace}
\newcommand{\qqp}{QQP\xspace}
\newcommand{\mrpc}{MRPC\xspace}
\newcommand{\mmlu}{MMLU\xspace}
\newcommand{\squad}{SQUAD V2\xspace}
\newcommand{\un}{UN Multi\xspace}
\newcommand{\iwslt}{IWSLT 2017\xspace}

\newcommand{\chat}{ChatGPT\xspace}
\newcommand{\llms}{LLMs\xspace}

\newcommand{\totalprompt}{$4,788$\xspace}
\newcommand{\totaldataset}{$13$\xspace}
\newcommand{\totaltask}{$8$\xspace}
\newcommand{\totalattack}{$7$\xspace}
\newcommand{\totalmodel}{$9$\xspace}

\begin{document}

\maketitle

\begin{abstract}
The increasing reliance on Large Language Models (LLMs) necessitates a comprehensive understanding of their robustness to prompts. 
In this paper, we introduce \method, a robustness benchmark designed to measure LLMs' resilience to adversarial prompts. 
This study uses a plethora of adversarial textual attacks on prompts across multiple levels: character, word, sentence, and semantic. 
The adversarial prompts, crafted to mimic plausible user errors like typos or synonyms, aim to evaluate how slight deviations can affect LLM outcomes while maintaining semantic integrity.
These prompts are then used in various tasks, including sentiment analysis, natural language inference, reading comprehension, machine translation, and math.
We generate \totalprompt adversarial prompts and evaluated over \totaltask tasks and \totaldataset datasets.
% , with $583,884$ test samples in total
Our findings demonstrate that LLMs are not robust to adversarial prompts. 
Furthermore, we present a comprehensive analysis to understand the mystery behind prompt robustness and its transferability.
We then offer insightful analysis and pragmatic recommendations for prompt composition, beneficial to both researchers and everyday users.
\end{abstract}

\vspace{-0.1in}
\section{Introduction}
\vspace{-0.05in}

\begin{wrapfigure}{r}{0.48\textwidth}
\vspace{-.2in}
    \includegraphics[width=.48\textwidth]{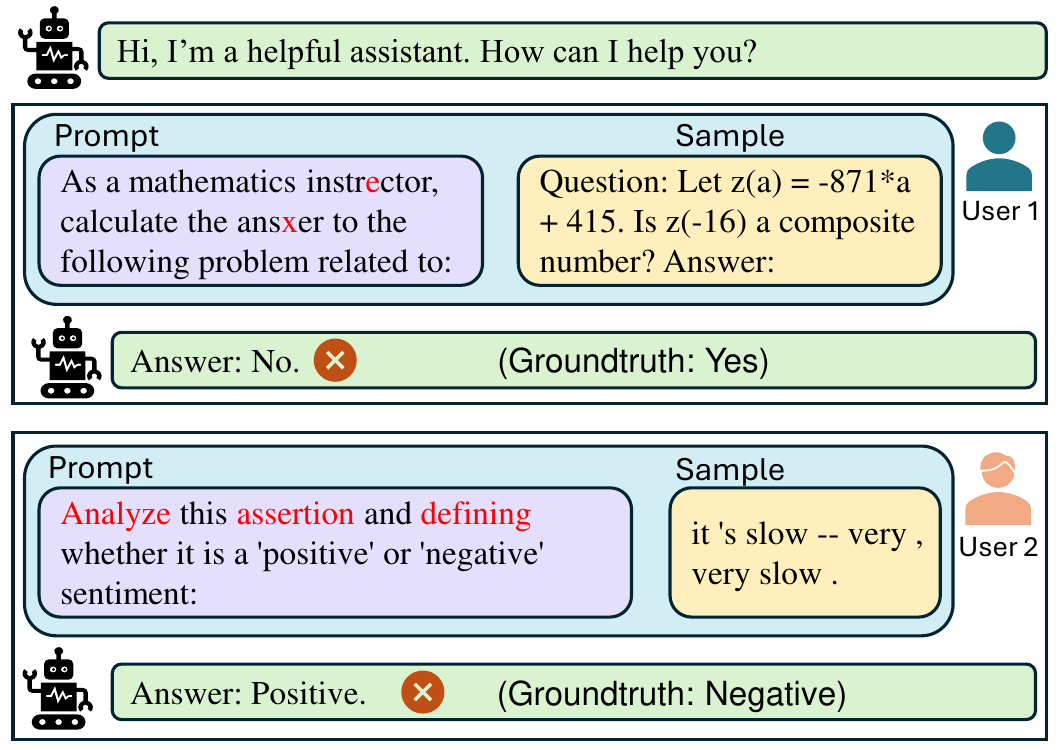}
    \vspace{-.2in}
    \caption{\llms are not robust to prompts: typos and synonyms lead to errors in math and sentiment analysis problems. The red characters and words are perturbations.}
    \label{fig-motiv}
\vspace{-.2in}
\end{wrapfigure}

Large language models (\llms) have gained increasing popularity due to their unprecedented performance in various tasks such as sentiment analysis~\citep{wang2019glue}, question answering~\citep{wang2019glue}, logical reasoning~\citep{liu2023evaluating}, etc.
An \emph{input} to an LLM is the concatenation of a \emph{prompt} and (optionally) a \emph{sample}, where the prompt aims to instruct the LLM what task to perform and the sample is the data for the task.
% Figure~\ref{fig-motiv} shows several examples of prompt, sample, and response when different users use \llms for different tasks. Note that a sample is optional in certain tasks. 
% For instance, in a task to write a country love story, a prompt  ``\prompt{Please write a story about country love}'' alone is sufficient. 

% \neil{Second para: Discuss the importance of robustness to input and existing works on robustness to sample perturbations.}

Given the popular adoption of \llms, particularly in the safety-critical and decision-making domains, it becomes essential to examine the \emph{robustness} of \llms to perturbations in an input. Indeed, existing work~\citep{wang2021adversarial,nie2020adversarial,wang2023robustness,zhuo2023robustness,yang2022glue} has attempted to assess the robustness of \llms from different perspectives.
For instance, AdvGLUE~\citep{wang2021adversarial} and ANLI~\citep{nie2020adversarial} are two public datasets to evaluate the robustness of language models to \emph{adversarial samples}, which are carefully perturbed samples to make a language model produce incorrect responses. 
In the era of large language models, \citet{wang2023robustness} evaluated \chat and other \llms with respect to their robustness to adversarial samples and out-of-distribution (OOD) samples. % using existing adversarial text datasets such as AdvGLUE~\citep{wang2021adversarial} and ANLI~\citep{nie2020adversarial}.
\citet{zhuo2023robustness} evaluated the robustness of \llms for a particular task called semantic parsing.
%Yang et al.~\citep{yang2022glue} evaluated OOD robustness by extending the GLUE \citep{wang2019glue} dataset.

 % \neil{Third para: Limitations of existing studies.}

 These studies demonstrated that the current \llms are not robust to adversarial and OOD samples for some popular natural language processing tasks.
 % However, in some application scenarios, an input only consists of a prompt without the need of a sample, making existing studies on robustness to adversarial samples not applicable. 
 A single prompt is often used to instruct an LLM to perform a task for multiple samples. For example, in a math problem (shown in Figure~\ref{fig-motiv}), a prompt can be used for multiple samples (i.e., math problems). Therefore, a perturbed prompt may cause an LLM to output incorrect responses for multiple clean samples. As a result, 
a perturbed prompt arguably has a larger impact on \llms than an adversarial sample, as the latter only influences the response of an LLM for a single sample. However, despite its central importance, the robustness of \llms to prompt perturbations is largely unexplored.

In this paper, we aim to bridge the gap by introducing \textbf{\method}, a comprehensive benchmark designed to evaluate the robustness of \llms to perturbations in prompts, understanding the factors that contribute to their robustness (or lack thereof), and identifying the key attributes of robust prompts. We consider a variety of prompt perturbations including 1) minor typos, synonyms, and different ways of expressing sentences with the same semantic meaning, which may commonly occur to normal users or developers in their daily use of \llms in \emph{non-adversarial settings}, as well as 2) perturbations strategically crafted by attackers in \emph{adversarial settings}. With a slight abuse of terminology, we call such a perturbed prompt in both scenarios \emph{adversarial prompt}. Figure~\ref{fig-motiv} shows examples of adversarial prompts with typos and synonyms that lead to incorrect responses.

% % Note that the term ``adversarial prompts'' here does not merely refer to attacks from malicious actors, but more importantly, stand out as \emph{simulations} to mimic potential typos, synonyms, or sentence styles that could possibly occur to daily LLM users.
% As shown in \figurename~\ref{fig-promptbench}, \method consists of \emph{prompts}, \emph{attacks}, \emph{models}, \emph{tasks}, \emph{datasets}, and \emph{analysis}.
% \method stands out with its ability to dynamically construct adversarial prompts based on the given datasets, models and clean prompts.
% The created adversarial prompts are then combined with input samples to generate adversarial inputs.\footnote{We focus on evaluating user prompts in this paper but our benchmark can also evaluate the robustness of system prompts. Moreover, we assume the samples are clean to better analyze the robustness of prompts.  Our benchmark can also be extended to analyze the robustness of  samples, as depicted in Sec.~\ref{sec-exp-advglue}.}
% This approach, contrasting with the prevalent use of static, pre-computed adversarial samples~\citep{nie2020adversarial, wang2021adversarial, wang2023robustness}, ensures a broader and more diverse set of adversarial inputs for each LLM.
% \method further supports different analysis to not only evaluate, but also understand the robustness of \llms from various perspectives.

\method consists of \emph{prompts}, \emph{attacks}, \emph{models}, \emph{tasks}, \emph{datasets}, and \emph{analysis}. 
%\method is a flexible benchmark that supports the popular open-source and proprietary models and datasets.
Specifically, we evaluate $4$ types of prompts: zero-shot (ZS), few-shot (FS), role-oriented, and task-oriented prompts.
We create $4$ types of attacks (called \emph{prompt attacks}) to craft adversarial prompts: \emph{character-level}, \emph{word-level}, \emph{sentence-level}, and \emph{semantic-level} attacks by extending \totalattack adversarial attacks~\citep{textbugger,deepwordbug,bertattack,textfooler,stresstest,checklist} that were originally designed to generate adversarial samples. 
We note that, although we call them attacks, their generated adversarial prompts %are not only attacks that may be performed by malicious actors to cause security concerns, but 
also serve as testbeds for \emph{mimicking} potential diverse prompts with naturally occurred perturbations from real LLM users. 
\method spans across \totalmodel prevalent LLMs, ranging from smaller models such as Flan-T5-large \citep{t5} to larger ones like ChatGPT \citep{chatgpt} and GPT-4 \citep{openai2023gpt4}. 
Moreover, we select \totaltask tasks for evaluation, namely, sentiment analysis (\sst~\citep{sst2}), grammar correctness (\cola~\citep{cola}), duplicate sentence detection (\qqp~\citep{qqp} and \mrpc~\citep{mrpc}), natural language inference (\mnli~\citep{mnli}, \qnli~\citep{wang2019glue}, \rte~\citep{wang2019glue}, and \wnli~\citep{wnli}), multi-task knowledge (MMLU~\citep{mmlu}), reading comprehension (SQuAD V2~\citep{squad}), translation (UN Multi~\citep{multiun} and IWSLT 2017~\citep{iwslt}), and math problem-solving (Mathematics~\citep{math}). 
In total, we created \totalprompt adversarial prompts, representing diverse, practical, and challenging scenarios.

We carry out extensive experiments and analysis using \method.
The results highlight a prevailing lack of robustness to adversarial prompts among current \llms, with word-level attacks proving the most effective ($39\%$ average performance drop in all tasks).
We delve into the reasons behind this vulnerability by exploring \llms' attention weights of each word in inputs for erroneous responses associated with clean and adversarial inputs, where an adversarial input is the concatenation of an adversarial prompt and a clean sample. 
Our findings reveal that adversarial prompts cause \llms to shift their focus towards the perturbed elements thus producing wrong responses.
We also examine the transferability of adversarial prompts between models, and suggest a successful transferability of adversarial prompts from one LLM to another.
Furthermore, we analyze word frequency patterns to guide future research in improving robustness and to aid end-users in crafting more robust prompts.
We conclude by discussing potential strategies for improving robustness.

To summarize, our contributions are as follows:
\begin{enumerate}[leftmargin=2em]
\setlength\itemsep{0em}

\item We introduce \method, the \emph{first} systematic benchmark for evaluating, understanding, and analyzing the robustness of \llms to adversarial prompts. 

% \item We conduct comprehensive evaluations of the robustness of LLMs to adversarial prompts, providing valuable benchmark results for the community.

% \item We extend adversarial attacks, which were originally designed to produce adversarial samples, to generate adversarial prompts. 

\item We conduct comprehensive evaluations on the robustness of LLMs to adversarial prompts and perform extensive analysis, including visual explanations for observed vulnerabilities, transferability analysis of adversarial prompts, and word frequency analysis to offer practical guidance to downstream users and prompt engineers to craft more robust prompts.

% \item In an effort to stimulate future research on LLMs' robustness, we also build a visualization website (Appendix~\ref{sec-append-vis}) to allow for easy exploration of adversarial prompts. We will make our code, compiled prompts, website, and evaluation benchmark available to the public. 
\end{enumerate}

% \begin{figure}[t!]
%     \centering
%     \includegraphics[width=.5\textwidth]{imgs/fig-promptbench.pdf}
%     \caption{The components of \method.}
%     \label{fig-promptbench}
% \end{figure}

\vspace{-0.1in}
\section{\method}
\label{sec-method}

% \method aims to provide a thorough evaluation of prompts on \llms.
% In this section, we introduce the basic modules of \method: prompts, models, tasks, datasets, attacks, and analysis.
\vspace{-0.05in}
\subsection{Prompts and models}
\label{sec-models}
\vspace{-0.05in}

We investigate four different types of prompt.
\textbf{Task-oriented} prompts explicitly describe the task the model is required to perform, which encourages the model to generate task-specific outputs based solely on its pre-training knowledge.
While \textbf{role-oriented} prompts typically frame the model as an entity with a specific role, such as an expert, advisor, or translator.
By incorporating role information, these prompts aim to implicitly convey the expected output format and behavior.
Each of the two categories of prompts can be designed for both \textbf{zero-shot (ZS)} and \textbf{few-shot (FS)} learning scenarios. In the zero-shot scenario, an input is defined as $[P, x]$, where $P$ denotes a prompt, $x$ is a sample, and $[,]$ denotes the concatenation operation. For the few-shot scenario, some examples are added to the input, resulting in the format $[P, E, x]$, where $E$ represents the examples. For instance, $E = \{[x_1, y_1], [x_2, y_2], [x_3, y_3]\}$ represents three examples in a three-shot learning scenario. In our experiments, we randomly select three examples in the training set of a task and append them to a prompt.
\cref{sec-append-prompt-example} shows examples of different types of prompts.

Our evaluation includes a diverse set of \llms to comprehensively assess their performance across various tasks and domains. The models we consider are as follows: Flan-T5-large \citep{t5} (0.8B), Dolly-6B \citep{dolly}, Vicuna-13B \citep{vicuna}, Llama2-13b-chat \citep{touvron2023llama2}, Cerebras-GPT-13B \citep{cerebras}, GPT-NEOX-20B \citep{neox}, Flan-UL2 (20B) \citep{ul2}, ChatGPT~\citep{chatgpt}, and GPT-4~\citep{openai2023gpt4}.\footnote{We did not perform prompt attacks on GPT-4 by optimizing the adversarial algorithms since it requires massive rounds of communications and is too costly. We used the adversarial prompts generated by \chat to evaluate GPT-4 since the adversarial prompts can be transferred (Sec. \ref{sec-transfer}).}
By incorporating \llms with different architectures and sizes, we aim to provide insights into their strengths and weaknesses, ultimately facilitating model selection for a specific application or use case.
Details of these \llms are in Appendix~\ref{sec-append-models}.
% Note that \method is flexible and supports all other \llms by extending the interface.

% \begin{figure}[t!]
%     \centering
%     \includegraphics[width=.45\textwidth]{imgs/fig-prompt.pdf}
%     \caption{Two possible interaction process by different users. User 1 inputs both the prompts and samples while user 2 only inputs the prompt. We assume that the system prompt is preset by the system and cannot be modified in this paper.}
%     \label{fig-prompt-motiv}
% \end{figure}
\vspace{-0.1in}
\subsection{Attacks}
\label{sec-attack}
\vspace{-0.05in}

% Multiple textual adversarial attacks were designed to generate adversarial samples~\citep{textbugger, textfooler, deepwordbug, bertattack, checklist, stresstest, zang2020word}. 
Given a single sample $x$ and its label $y$, a textual adversarial attack aims to find a perturbation $\delta$ such that an LLM $f_\theta$ produces an incorrect response. Formally, $\delta$ is found by solving the following optimization problem: $\mathop{\max}_{\delta \in \mathcal{C}} \mathcal{L} [f_\theta(x+\delta); y]$, where $x+\delta$ is the adversarial sample, $f_\theta(x+\delta)$ is the response of the LLM when taking the adversarial sample alone as input, $\mathcal{C}$ indicates the constraints for the perturbation $\delta$, and $\mathcal{L}$ represents a loss function.

% attack an LLM $f_\theta$ by perturbing each sample $x$ with $\delta$ given certain budget $\mathcal{C}$: $\mathop{\arg \max}_{\delta \in \mathcal{C}} \mathcal{L} [f_\theta(x+\delta), y]$ to maximize the loss on the sample $x$, where $\mathcal{L}$ represents the loss function.

\textbf{Prompt attack.}
In this paper, our focus is to attack the \emph{prompts} rather than samples. 
This is due to the popularity of \llms in different applications, which generate responses using in-context learning on prompts (i.e., instructions) and samples.
Prompts are either input by users or generated by the system or developers.
% While the system prompts are often preset by service providers for proprietary models, user prompts are often generated by daily users, which is our focus.
Moreover, the ultimate purpose of performing such ``attack'' is actually \emph{not} to genuinely attack the models, but to \emph{simulate} possible perturbations that may naturally occur in real situations.
\tablename~\ref{table: adv prompt examples} shows multiple prompts generated by adversarial approaches that are used to mimic possible user prompts, which are popular errors or expressions made by users.
Since users can make different mistakes when entering prompts, such as typos, different word usage, different sentence styles, etc., the study on prompt robustness is necessary to understand \llms.

We denote an input to \llms as $[P, x]$, where $P$ is a prompt, $x$ is a sample, and $[,]$ denotes concatenation. Note that in the few-shot learning scenario, a few examples are appended to the prompt; and the sample $x$ is optional in certain application scenarios. Our prompt attack can also be extended to such scenarios, but we use the notation $[P, x]$ for simplicity. Given a dataset $\mathcal{D} = \{ (x_i, y_i)\}_{i \in [N]}$ with $N$ samples and their ground-truth labels, a prompt attack aims to perturb $P$ such that an LLM $f_\theta$ produces incorrect responses for all samples in the dataset $\mathcal{D}$. 

\begin{definition}[Prompt Attack]
Given an LLM $f_\theta$, a dataset $\mathcal{D} = \{ (x_i, y_i)\}_{i \in [N]}$, and a clean prompt $P$, the objective of a prompt attack can be formulated as follows:
\begin{align}
\label{eq-promptattack}
\mathop{\max}_{\delta \in \mathcal{C}} \sum_{(x; y) \in \mathcal{D}} \mathcal{L} [f_\theta([P+\delta, x]), y],
\end{align}
where $\delta$ is the textual perturbation added to the clean prompt $P$ and $\mathcal{C}$ is the allowable perturbation set, i.e., perturbation constraint. We note that this attack is analogous to universal adversarial perturbation (UAP) \citep{uap, brown2017adversarial} and universal adversarial trigger (UAT) \citep{uat}, extending these concepts to the realm of prompts. \cref{sec-append-textual} shows more comparisons.
\end{definition}

\textbf{Different attacks.}
We then modify the existing black-box textual attacks to implement Eq.~(\ref{eq-promptattack}) due to their efficiency and no reliance on the model gradient.
Thus, both open-sourced and proprietary LLMs can be attack targets.
Our instantiations span four distinct levels, capturing a broad spectrum of complexities from simple character manipulations to sophisticated semantic alterations. Details of each attack are in \cref{sec-append-attacks}.

\begin{itemize}
\setlength\itemsep{0em}
    \item \textbf{Character-level:} We employ TextBugger~\citep{textbugger} and DeepWordBug~\citep{deepwordbug}, which manipulate texts by introducing typos or errors to words, e.g., by adding, deleting, repeating, replacing, and permuting characters for certain words.
    \item \textbf{Word-level:} We use BertAttack~\citep{bertattack} and TextFooler~\citep{textfooler} to replace words with synonyms or contextually similar words to deceive \llms.
    \item \textbf{Sentence-level:} 
    We implement StressTest~\citep{stresstest} and CheckList~\citep{checklist} to append irrelevant or extraneous sentences to the end of prompts, intending to distract \llms.
    For StressTest, we adopt settings similar to those in \citep{wang2019glue}, appending ``\prompt{and true is true}'', ``\prompt{and false is not true}'', or ``\prompt{and true is true}'' for five times to the end of a prompt. For the CheckList attack, we generate $50$ random sequences consisting of alphabets and digits, each with a length of $10$, and append this random sequence to the end of a prompt.
    % For instance, in the CheckList attack, we generate $50$ random sequences of alphabets and digits, then randomly select one sequence and append it at the end of the input.
    \item \textbf{Semantic-level:} We simulate the linguistic behavior of people from different countries by choosing $6$ common languages (Chinese, French, Arabic, Spanish, Japanese, and Korean) and constructing $10$ prompts for each language per dataset. These prompts are then translated into English, introducing linguistic nuances and variations that could potentially impact LLMs.
\end{itemize}

\textbf{Semantic preservation of adversarial prompts.}

% \usepackage{multirow}
\begin{wraptable}{r}{7.5cm}
% \begin{table}
\vspace{-0.3in}
\caption{Statistics of datasets used in this paper.}
\label{table: task summary}
\centering 
\resizebox{.52\textwidth}{!}{
\begin{tabular}{lcccc}
\toprule
Task                                          & Dataset    & \#Sample & \#Class & \#[Adv. prompt, sample]\\
\midrule
Sentiment analysis                            & SST2       & 872  & 2 & 73,248\\
\midrule
Grammar correctness                           & CoLA       & 1,000 & 2  & 84,000\\
\midrule
\multirow{2}{*}{\makecell[l]{Duplicate sentence\\detection}} & QQP        & 1,000 & 2 & 84,000\\
                                              & MRPC       & 408  & 2 & 34,272\\ 
\midrule                                              
\multirow{4}{*}{\makecell[l]{Natural language\\inference}}   & MNLI       & 1,000 & 3 & 84,000\\
                                              & QNLI       & 1,000 & 2 & 84,000\\
                                              & RTE        & 277  & 2 & 23,268\\
                                              & WNLI       & 71   & 2 & 5,964\\
\midrule
Multi-task knowledge                          & MMLU       & 564  & 4 & 47,376\\
\midrule

Reading comprehension                         & SQuAD V2   & 200  & - & 16,800\\
\midrule
\multirow{2}{*}{Translation}                  & Multi UN   & 99   & - & 8,316\\
                                              & IWSLT 2017 & 100  & - & 8,400\\
\midrule
Math reasoning                                      & Math       & 160  & - & 13,440
% \\
% \midrule
% Logical reasoning & Boolean Expression & 100 & 2 & 8400 \\
% \midrule
% Algorithm & Valid Parentheses & 100 & 2 & 8400
\\
\bottomrule
\end{tabular}
}
% \end{table}
\end{wraptable}

Are adversarial prompts realistic?
Our purpose is to simulate plausible user errors; thus, it is imperative that these prompts preserve semantic integrity, ensuring they remain both \emph{acceptable} and \emph{imperceptible} to human comprehension.
It is of paramount importance that our adversarially engineered prompts retain coherence and realism.
Our human study in \cref{sec-append-semantic} shows that at least $85\%$ of subjects agreed that the generated prompts are acceptable.

\vspace{-0.1in}
\subsection{Tasks and datasets}
\label{sec-task-dataset}
\vspace{-0.05in}

Currently, \method supports \totaltask tasks and \totaldataset datasets across sentiment analysis to math reasoning.
Due to space limit, we leave the details of the datasets in \cref{appendix: dataset and models}.

\vspace{-0.1in}
\section{Experiments}
\vspace{-0.1in}

\textbf{Setup.}
The extensive computational need to generate an adversarial prompt requires iterating throughout the dataset $100$ on average.
Therefore, the evaluation of an entire dataset using \llms is unfeasible.
To alleviate the computational constraint and preserve a fair study process, we adopt a sampling strategy that entails selecting a subset of samples from the validation or test sets across various datasets.
The statistics of each dataset and tasks are summarized in \tablename~\ref{table: task summary}.\footnote{In \tablename~\ref{table: task summary}, the last column denotes the total evaluation sample size for each dataset on each model. For instance, there are $872$ test samples in SST-2 dataset and each sample should go through $7$ adversarial attacks on $4$ types of prompts, each with $3$ prompts, thus the test size for each model is $872 \times 7 \times 4 \times 3 =73248$.}
The sampling details are in \cref{sec-append-exp}.

We initially assess the performance of all LLMs without prompt attacks to provide a performance baseline. We find that certain LLMs do not even demonstrate satisfactory performance with clean prompts, narrowing our selection to $6$ \llms: Flan-T5-large, Vicuna-13B, Llama2-13B-chat, UL2, \chat, and GPT-4.
Further details and discussions on clean prompt performance in all LLMs are available in Appendix~\ref{sec-append-exp-cleanresults}.
We generate $10$ distinct prompts for both role-oriented and task-oriented categories.
Each prompt can be augmented with three examples, forming the few-shot prompts.
In total, we have $40$ prompts for each dataset on each LLM.
For better efficiency and performance, we select the top $3$ best-performing prompts of each type to perform prompt attacks.
As a result, we evaluate the adversarial vulnerabilities of \totalmodel \llms across \totaldataset datasets, encompassing a total of \totalprompt prompts\footnote{\totalprompt$=3 \times 4 \times 5 \times 13 \times 7 - 336 \times 3 \times 2$, where the numbers on the R.H.S. denote \#attacked prompts, \#prompt types, \#\llms, \#datasets, and \#attacks, respectively. We did not conduct attacks on Vicuna, Llama2-13B-chat on certain datasets because the output of these datasets are meaningless, so that we subtract $336 \times 3 \times 2$ prompts.} and their respective adversarial counterparts. This comprehensive evaluation allows us to gain valuable insight into the robustness and performance of \llms in a wide range of scenarios and prompt styles.

\begin{table*}[t!]
\centering
\caption{The APDR and standard deviations of different attacks on different datasets.}
% \vspace{-.1in}
\label{tb-dataset-attack}
\resizebox{.8\textwidth}{!}{
\begin{tabular}{lccccccc}
\toprule

\multirow{2}{*}{Dataset} & \multicolumn{2}{c}{Character-level} & \multicolumn{2}{c}{Word-level} & \multicolumn{2}{c}{Sentence-level} & \multicolumn{1}{c}{Semantic-level} \\
\cmidrule(lr){2-3} \cmidrule(lr){4-5} \cmidrule(lr){6-7} \cmidrule(lr){8-8}
& TextBugger & DeepWordBug & TextFooler & BertAttack & CheckList & StressTest & Semantic \\
\midrule
SST-2    & 0.25\scriptsize{$\pm$0.39} & 0.18\scriptsize{$\pm$0.33} & 0.35\scriptsize{$\pm$0.41} & 0.34\scriptsize{$\pm$0.44} & 0.22\scriptsize{$\pm$0.36} & 0.15\scriptsize{$\pm$0.31} & 0.28\scriptsize{$\pm$0.35} \\
CoLA     & 0.39\scriptsize{$\pm$0.40} & 0.27\scriptsize{$\pm$0.32} & 0.43\scriptsize{$\pm$0.35} & 0.45\scriptsize{$\pm$0.38} & 0.23\scriptsize{$\pm$0.30} & 0.18\scriptsize{$\pm$0.25} & 0.34\scriptsize{$\pm$0.37} \\
QQP      & 0.30\scriptsize{$\pm$0.38} & 0.22\scriptsize{$\pm$0.31} & 0.31\scriptsize{$\pm$0.36} & 0.33\scriptsize{$\pm$0.38} & 0.18\scriptsize{$\pm$0.30} & 0.06\scriptsize{$\pm$0.26} & 0.40\scriptsize{$\pm$0.39} \\
MRPC     & 0.37\scriptsize{$\pm$0.42} & 0.34\scriptsize{$\pm$0.41} & 0.37\scriptsize{$\pm$0.41} & 0.42\scriptsize{$\pm$0.38} & 0.24\scriptsize{$\pm$0.37} & 0.25\scriptsize{$\pm$0.33} & 0.39\scriptsize{$\pm$0.39} \\
MNLI     & 0.32\scriptsize{$\pm$0.40} & 0.18\scriptsize{$\pm$0.29} & 0.32\scriptsize{$\pm$0.39} & 0.34\scriptsize{$\pm$0.36} & 0.14\scriptsize{$\pm$0.24} & 0.10\scriptsize{$\pm$0.25} & 0.22\scriptsize{$\pm$0.24} \\
QNLI     & 0.38\scriptsize{$\pm$0.39} & 0.40\scriptsize{$\pm$0.35} & 0.50\scriptsize{$\pm$0.39} & 0.52\scriptsize{$\pm$0.38} & 0.25\scriptsize{$\pm$0.39} & 0.23\scriptsize{$\pm$0.33} & 0.40\scriptsize{$\pm$0.35} \\
RTE      & 0.33\scriptsize{$\pm$0.41} & 0.25\scriptsize{$\pm$0.35} & 0.37\scriptsize{$\pm$0.44} & 0.40\scriptsize{$\pm$0.42} & 0.18\scriptsize{$\pm$0.32} & 0.17\scriptsize{$\pm$0.24} & 0.42\scriptsize{$\pm$0.40} \\
WNLI     & 0.39\scriptsize{$\pm$0.42} & 0.31\scriptsize{$\pm$0.37} & 0.41\scriptsize{$\pm$0.43} & 0.41\scriptsize{$\pm$0.40} & 0.24\scriptsize{$\pm$0.32} & 0.20\scriptsize{$\pm$0.27} & 0.49\scriptsize{$\pm$0.39} \\
MMLU     & 0.21\scriptsize{$\pm$0.24} & 0.12\scriptsize{$\pm$0.16} & 0.21\scriptsize{$\pm$0.20} & 0.40\scriptsize{$\pm$0.30} & 0.13\scriptsize{$\pm$0.18} & 0.03\scriptsize{$\pm$0.15} & 0.20\scriptsize{$\pm$0.19} \\
SQuAD V2 & 0.09\scriptsize{$\pm$0.17} & 0.05\scriptsize{$\pm$0.08} & 0.25\scriptsize{$\pm$0.29} & 0.31\scriptsize{$\pm$0.32} & 0.02\scriptsize{$\pm$0.03} & 0.02\scriptsize{$\pm$0.04} & 0.08\scriptsize{$\pm$0.09} \\
IWSLT    & 0.08\scriptsize{$\pm$0.14} & 0.10\scriptsize{$\pm$0.12} & 0.27\scriptsize{$\pm$0.30} & 0.12\scriptsize{$\pm$0.18} & 0.10\scriptsize{$\pm$0.10} & 0.17\scriptsize{$\pm$0.19} & 0.18\scriptsize{$\pm$0.14} \\
UN Multi & 0.06\scriptsize{$\pm$0.08} & 0.08\scriptsize{$\pm$0.12} & 0.15\scriptsize{$\pm$0.19} & 0.10\scriptsize{$\pm$0.16} & 0.06\scriptsize{$\pm$0.07} & 0.09\scriptsize{$\pm$0.11} & 0.15\scriptsize{$\pm$0.18} \\
Math     & 0.18\scriptsize{$\pm$0.17} & 0.14\scriptsize{$\pm$0.13} & 0.49\scriptsize{$\pm$0.36} & 0.42\scriptsize{$\pm$0.32} & 0.15\scriptsize{$\pm$0.11} & 0.13\scriptsize{$\pm$0.08} & 0.23\scriptsize{$\pm$0.13} \\
\midrule
Avg      & 0.21\scriptsize{$\pm$0.30} & 0.17\scriptsize{$\pm$0.26} & 0.31\scriptsize{$\pm$0.33} & 0.33\scriptsize{$\pm$0.34} & 0.12\scriptsize{$\pm$0.23} & 0.11\scriptsize{$\pm$0.23}  & 0.22\scriptsize{$\pm$0.26}
\\
\bottomrule

\end{tabular}
}
\vspace{-0.2in}
\end{table*}

\begin{table*}[t!]
\parbox{.55\linewidth}
{
% \vspace{13pt}
\caption{The APDR on different \llms.
% (T5=Flan-T5-large, Vicuna=Vicuna-13B, UL2=Flan-UL2).
}
\label{tb-dataset-model}
\centering
%\begin{center}
\resizebox{.55\textwidth}{!}{
\begin{tabular}{lcccccc}
\toprule

Dataset   & T5-large         & Vicuna     & Llama2 & UL2        & \chat & GPT-4  \\ \midrule

SST-2    & 0.04\scriptsize{$\pm$0.11} & 0.83\scriptsize{$\pm$0.26} & 0.24\scriptsize{$\pm$0.33} & 0.03\scriptsize{$\pm$0.12} & 0.17\scriptsize{$\pm$0.29} & 0.24\scriptsize{$\pm$0.38} \\
CoLA     & 0.16\scriptsize{$\pm$0.19} & 0.81\scriptsize{$\pm$0.22} & 0.38\scriptsize{$\pm$0.32} & 0.13\scriptsize{$\pm$0.20} & 0.21\scriptsize{$\pm$0.31} & 0.13\scriptsize{$\pm$0.23} \\
QQP      & 0.09\scriptsize{$\pm$0.15} & 0.51\scriptsize{$\pm$0.41} & 0.59\scriptsize{$\pm$0.33} & 0.02\scriptsize{$\pm$0.04} & 0.16\scriptsize{$\pm$0.30} & 0.16\scriptsize{$\pm$0.38} \\
MRPC     & 0.17\scriptsize{$\pm$0.26} & 0.52\scriptsize{$\pm$0.40} & 0.84\scriptsize{$\pm$0.27} & 0.06\scriptsize{$\pm$0.10} & 0.22\scriptsize{$\pm$0.29} & 0.04\scriptsize{$\pm$0.06} \\
MNLI     & 0.08\scriptsize{$\pm$0.13} & 0.67\scriptsize{$\pm$0.38} & 0.32\scriptsize{$\pm$0.32} & 0.06\scriptsize{$\pm$0.12} & 0.13\scriptsize{$\pm$0.18} & -0.03\scriptsize{$\pm$0.02} \\
QNLI     & 0.33\scriptsize{$\pm$0.25} & 0.87\scriptsize{$\pm$0.19} & 0.51\scriptsize{$\pm$0.39} & 0.05\scriptsize{$\pm$0.11} & 0.25\scriptsize{$\pm$0.31} & 0.05\scriptsize{$\pm$0.23} \\
RTE      & 0.08\scriptsize{$\pm$0.13} & 0.78\scriptsize{$\pm$0.23} & 0.68\scriptsize{$\pm$0.39} & 0.02\scriptsize{$\pm$0.04} & 0.09\scriptsize{$\pm$0.13} & 0.03\scriptsize{$\pm$0.05} \\
WNLI     & 0.13\scriptsize{$\pm$0.14} & 0.78\scriptsize{$\pm$0.27} & 0.73\scriptsize{$\pm$0.37} & 0.04\scriptsize{$\pm$0.03} & 0.14\scriptsize{$\pm$0.12} & 0.04\scriptsize{$\pm$0.04} \\
MMLU     & 0.11\scriptsize{$\pm$0.18} & 0.41\scriptsize{$\pm$0.24} & 0.28\scriptsize{$\pm$0.24} & 0.05\scriptsize{$\pm$0.11} & 0.14\scriptsize{$\pm$0.18} & 0.04\scriptsize{$\pm$0.04} \\
SQuAD V2 & 0.05\scriptsize{$\pm$0.12} & -                          & -                          & 0.10\scriptsize{$\pm$0.18} & 0.22\scriptsize{$\pm$0.28} & 0.27\scriptsize{$\pm$0.31} \\
IWSLT    & 0.14\scriptsize{$\pm$0.17} & -                          & -                          & 0.15\scriptsize{$\pm$0.11} & 0.17\scriptsize{$\pm$0.26} & 0.07\scriptsize{$\pm$0.14} \\
UN Multi & 0.13\scriptsize{$\pm$0.14} & -                          & -                          & 0.05\scriptsize{$\pm$0.05} & 0.12\scriptsize{$\pm$0.18} & -0.02\scriptsize{$\pm$0.01} \\
Math     & 0.24\scriptsize{$\pm$0.21} & -                          & -                          & 0.21\scriptsize{$\pm$0.21} & 0.33\scriptsize{$\pm$0.31} & 0.02\scriptsize{$\pm$0.18} \\
\midrule
Avg      & 0.13\scriptsize{$\pm$0.19} & 0.69\scriptsize{$\pm$0.34} & 0.51\scriptsize{$\pm$0.39} & 0.08\scriptsize{$\pm$0.14} & 0.18\scriptsize{$\pm$0.26} & 0.08\scriptsize{$\pm$0.21}\\
\bottomrule
\end{tabular}
%\end{center}
}
}
\hfill
{
\parbox{.4\linewidth}{

\caption{APDR on different prompts.
% (ZS=zero-shot, FS=few-shot, ``-task'' and ``-role'' denote task- and role-oriented prompts).
}
\label{tb-dataset-prompt}
\centering
%\begin{center}
\resizebox{.4\textwidth}{!}{
\begin{tabular}{lcccc}
\toprule
Dataset             & ZS-task & ZS-role & FS-task & FS-role \\ \midrule
SST-2    & 0.31\scriptsize{$\pm$0.39} & 0.28\scriptsize{$\pm$0.35} & 0.22\scriptsize{$\pm$0.38} & 0.24\scriptsize{$\pm$0.39} \\
CoLA     & 0.43\scriptsize{$\pm$0.35} & 0.43\scriptsize{$\pm$0.38} & 0.24\scriptsize{$\pm$0.28} & 0.25\scriptsize{$\pm$0.36} \\
QQP      & 0.43\scriptsize{$\pm$0.42} & 0.34\scriptsize{$\pm$0.43} & 0.16\scriptsize{$\pm$0.21} & 0.14\scriptsize{$\pm$0.20} \\
MRPC     & 0.44\scriptsize{$\pm$0.44} & 0.51\scriptsize{$\pm$0.43} & 0.24\scriptsize{$\pm$0.32} & 0.23\scriptsize{$\pm$0.30} \\
MNLI     & 0.29\scriptsize{$\pm$0.35} & 0.26\scriptsize{$\pm$0.33} & 0.19\scriptsize{$\pm$0.29} & 0.21\scriptsize{$\pm$0.33} \\
QNLI     & 0.46\scriptsize{$\pm$0.39} & 0.51\scriptsize{$\pm$0.40} & 0.30\scriptsize{$\pm$0.34} & 0.32\scriptsize{$\pm$0.36} \\
RTE      & 0.33\scriptsize{$\pm$0.39} & 0.35\scriptsize{$\pm$0.40} & 0.31\scriptsize{$\pm$0.39} & 0.27\scriptsize{$\pm$0.38} \\
WNLI     & 0.36\scriptsize{$\pm$0.36} & 0.39\scriptsize{$\pm$0.39} & 0.37\scriptsize{$\pm$0.41} & 0.33\scriptsize{$\pm$0.38} \\
MMLU     & 0.25\scriptsize{$\pm$0.23} & 0.22\scriptsize{$\pm$0.26} & 0.18\scriptsize{$\pm$0.23} & 0.14\scriptsize{$\pm$0.20} \\
SQuAD V2 & 0.16\scriptsize{$\pm$0.26} & 0.20\scriptsize{$\pm$0.28} & 0.06\scriptsize{$\pm$0.11} & 0.07\scriptsize{$\pm$0.12} \\
IWSLT    & 0.18\scriptsize{$\pm$0.22} & 0.24\scriptsize{$\pm$0.25} & 0.08\scriptsize{$\pm$0.09} & 0.11\scriptsize{$\pm$0.10} \\
UN Multi & 0.17\scriptsize{$\pm$0.18} & 0.15\scriptsize{$\pm$0.16} & 0.04\scriptsize{$\pm$0.07} & 0.04\scriptsize{$\pm$0.07} \\
Math     & 0.33\scriptsize{$\pm$0.26} & 0.39\scriptsize{$\pm$0.30} & 0.16\scriptsize{$\pm$0.18} & 0.17\scriptsize{$\pm$0.17} \\
\midrule
Avg      & 0.33\scriptsize{$\pm$0.36} & 0.34\scriptsize{$\pm$0.37} & 0.21\scriptsize{$\pm$0.31} & 0.21\scriptsize{$\pm$0.31} 
\\ \bottomrule 
\end{tabular}
}
}
}
%\end{center}
\vspace{-.2in}
\end{table*}

\textbf{Evaluation metrics.}
Considering the diverse evaluation metrics across tasks and varying baseline performances across models and datasets, the absolute performance drop may not provide a meaningful comparison. Thus, we introduce a unified metric, the \emph{Performance Drop Rate} (PDR). PDR quantifies the relative performance decline following a prompt attack, offering a contextually normalized measure to compare different attacks, datasets, and models.
The PDR is given by: $\mathit{PDR}(A, P, f_\theta, \mathcal{D}) = 1 - \frac{\sum_{(x;y) \in \mathcal{D}} { \mathcal{M} [ f_{\theta}([A(P), x]), y]}}{ \sum_{(x;y) \in \mathcal{D}} {\mathcal{M} [f_{\theta}([P, x]), y]}},$
where $A$ is the adversarial attack applied to prompt $P$, and $\mathcal{M}[\cdot]$ is the evaluation function: for classification task, $\mathcal{M}[\cdot]$ is the indicator function $\mathds{1}[\hat{y}, y]$ which equals to $1$ when $\hat{y} = y$, and $0$ otherwise; for reading comprehension task, $\mathcal{M}[\cdot]$ is the F1-score; for translation tasks, $\mathcal{M}[\cdot]$ is the Bleu metric \citep{bleu}. Note that a negative PDR implies that adversarial prompts can enhance performance.

\vspace{-0.1in}
\section{Results and analysis}
\label{sec-attacksection}
% \vspace{-0.1in}
% In this section, we present our benchmark results and analysis in evaluating the robustness of \llms on adversarial prompts.
\vspace{-0.1in}
\subsection{Results across attacks, models, and prompts}
\vspace{-0.05in}

We report and discuss the Average PDR (APDR) across different attacks, \llms, and prompts.
Our main results are based on \textit{all} the prompts, whose conclusions are in consistent with Appendix~\ref{appendix-excluded-results} where we show result by excluding unacceptable adversarial prompts.
Note that although our semantic preserving study demonstrated that at least 85\% of adversarial prompts are acceptable, there are still some adversarial prompts diverged from their intended semantic meaning.
Furthermore, note that the discrepancies in APDR variance values are due to varying PDR values across different attacks, prompts, models and datasets.

% The results of Llama2-13B-chat, Vicuna-13B-v1.3 are shown in Appendix~\ref{appendix-new-llms}. The results of Boolean Expression dataset and Valid Parentheses dataset are shown in Appendix~\ref{appendix-new-datasets}.

% In this subsection, we examine the ADR of different adversarial attacks on different datasets to gain a better understanding of their vulnerabilities and identify the most effective attack strategies. Our analysis of the ADR for various attacks on different LLMs is presented in Table 2.

\textbf{Analysis on attacks.}
\tablename~\ref{tb-dataset-attack} summarizes the APDR of \totalattack attacks on \totaldataset datasets, calculated by
    $\textit{APDR}_{A}(A, \mathcal{D}) = \frac{1}{|\mathcal{P}|} \frac{1}{|\mathcal{F}|} \sum_{P \in \mathcal{P}} \sum_{f_\theta \in \mathcal{F}} \mathit{PDR}(A, P, f_\theta, \mathcal{D})$,
where $\mathcal{P}$ is the set of $4$ types of prompts and $\mathcal{F}$ is the set of all models. The results offer several key insights. Firstly, attack effectiveness is highly variable, with word-level attacks proving the most potent, leading to an average performance decline of $33\%$ across \textit{all} datasets.
Character-level attacks rank second, inducing a $20\%$ performance drop in most datasets.
Notably, semantic-level attacks exhibit potency nearly commensurate with character-level attacks, emphasizing the profound impact of nuanced linguistic variations on \llms' performance. On the contrary, sentence-level attacks pose less of a threat, suggesting adversarial interventions at this level have a diminished effect. Moreover, the effect of prompt attack varies across different datasets. For instance, StressTest attacks on \squad yield a mere 2\% performance drop, while inflicting a 25\% drop on MRPC. Furthermore, we observe that the StressTest attack paradoxically bolsters the model's performance in some datasets, we delve into this phenomenon in \cref{stresstest analysis}.

Note that while character-level attacks are detectable by grammar detection tools, word- and semantic-level attacks underscore the importance of robust semantic understanding and accurate task presentation/translation for \llms. A comprehensive understanding of these nuances will inform a deeper comprehension of adversarial attacks on \llms.

\textbf{Analysis on \llms.}
\tablename~\ref{tb-dataset-model} summarizes the APDR of \totalmodel \llms on \totaldataset datasets, calculated by
$
    \mathit{APDR}_{f_\theta}(f_\theta, \mathcal{D}) = \frac{1}{|\mathcal{A}|} \frac{1}{|\mathcal{P}|} \sum_{ A \in \mathcal{A}} \sum_{P \in \mathcal{P}} \mathit{PDR}(A, P, f_\theta, \mathcal{D}),
$
where $\mathcal{P}$ is the set of $4$ types of prompts and $\mathcal{A}$ is the set of \totalattack attacks.
Our analysis reveals that GPT-4 and UL2 significantly outperform other models in terms of robustness, followed by T5-large, \chat, and Llama2, with Vicuna presenting the least robustness. The robustness against adversarial prompts of UL2, T5-large, and \chat varies across datasets, with UL2 and T5-large showing less vulnerability to attacks on sentiment classification (SST-2), most NLI tasks, and reading comprehension (SQuAD V2). Specifically, UL2 excels in translation tasks, while \chat displays robustness in certain NLI tasks. Vicuna, however, exhibits consistently high susceptibility to attacks across all tasks. 
It can be seen that, given the same adversarial prompts generated by 
\chat, GPT-4 exhibits superior robustness in all tasks. However, it is crucial to realize that this observed robustness might attribute to the weak transferability of the adversarial prompts crafted specifically for \chat.
In the future, the performance of GPT-4 and ChatGPT could be improved since these proprietary models keep evolving.
\textbf{Models vs. attacks.} We show in \tablename~\ref{tb-model-attack} the relation between models and attacks.
Generally, word-level attacks emerge as the most potent, and BertAttack consistently outperforms others across all models. However, no discernible pattern emerges for the efficacy of the other attacks. For instance, while TextBugger proves more effective than DeepWordBug for some models such as Llama2 and ChatGPT, the inverse holds true for T5-large. Notably, Vicuna and Llama2 are distinctly vulnerable to sentence-level attacks, in contrast to models like T5-large and \chat, which remain largely unaffected. Such observations may hint at inherent vulnerabilities specific to Llama-based models.

\textbf{Analysis on types of prompts.}
\tablename~\ref{tb-dataset-prompt} summarizes the APDR of $4$ types of prompts on \totaldataset datasets, calculated by 
$
    \textit{APDR}_{t}(\mathcal{D}) = \frac{1}{|\mathcal{A}|} \frac{1}{|\mathcal{P}_{t}|} \frac{1}{|\mathcal{F}|} \sum_{\mathcal{A} \in \mathcal{A}} \sum_{P \in \mathcal{P}_t} \sum_{f_\theta \in \mathcal{F}} \mathit{PDR}(A, P, f_\theta, \mathcal{D}),
$
where $\mathcal{P}_{t}$ is the set of prompts of certain type $t$, $\mathcal{A}$ is the set of \totalattack attacks and $\mathcal{F}$ is the set of all models. In our analysis, few-shot prompts consistently demonstrate superior robustness compared to zero-shot prompts across all datasets. 
Furthermore, while task-oriented prompts marginally outperform role-oriented prompts in overall robustness, both show varying strengths across different datasets and tasks. 
For example, role-oriented prompts present increased robustness within the SST-2 and QQP datasets, while task-oriented prompts are more resilient within the MRPC, QNLI, SQuAD V2, and IWSLT datasets.
Insights into different effects of prompt types on model vulnerability can inform better prompt design and tuning strategies, enhancing \llms robustness against adversarial prompt attacks.

\vspace{-0.1in}
\subsection{Results on model size, fine-tuning, and adversarial inputs}
\vspace{-0.05in}

\textbf{Model size and fine-tuning.} We analyze the performance on different models sizes using the Llama2 series (7B, 13B, and 70B).
Our results in \cref{sec-append-size-finetune} show that larger models are typically more robust than smaller ones, but exceptions can occur when smaller models outperform larger ones, which is an interesting finding that can trigger future research.
We also evaluated the impact of fine-tuning using vanilla Llama2 and Llama2-chat models in \cref{sec-append-size-finetune}, indicating that fine-tuned models are generally better at adversarial prompts.

\textbf{Attacking both prompts and samples.}
We attacked both prompts and tested on AdvGLUE~\citep{wang2021adversarial}, which contains adversarial samples.
Our results in \cref{tb-advglue-all} show that attacking both will perform even worse.
However, intriguing things happen, since attacking both can sometimes enhance the performance that needs further effort.

\begin{table*}[t!]
\caption{Attention visualization of samples that are \textit{correctly classified by clean prompts but misclassified by adv. prompts}. For each attack, the above is the \textit{clean prompt} with sample text, the below is the corresponding \textit{adversarial prompt} with the same sample text. N=Negative, P=Positive and N/A means the response is not available. The {\color{green}{green}} and {\color{red}{red}} color denote right and wrong answers, respectively. Color intensity denotes different attention weights (heavier color means larger weights).}
\label{table: attention misclassified adv prompts}
\centering
\resizebox{\textwidth}{!}{
\begin{tabular}{l c m{18.5cm}}
\toprule
\multicolumn{1}{c}{\textbf{Attack}} & \multicolumn{1}{c}{\textbf{Pred.}} & \multicolumn{1}{c}{\textbf{[Prompt, sample]}} \\
\midrule
\multirow{5}{*}{\footnotesize BertAttack} &  {\footnotesize \color{green}{N}} & \scriptsize \colorbox[RGB]{254,233,223}{In\vphantom{fg}}\hspace*{0pt}\colorbox[RGB]{254,233,224}{the\vphantom{fg}}\hspace*{0pt}\colorbox[RGB]{254,229,217}{role\vphantom{fg}}\hspace*{0pt}\colorbox[RGB]{252,189,163}{of\vphantom{fg}}\hspace*{0pt}\colorbox[RGB]{254,229,218}{a\vphantom{fg}}\hspace*{0pt}\colorbox[RGB]{251,132,100}{sentiment\vphantom{fg}}\hspace*{0pt}\colorbox[RGB]{253,215,199}{analysis\vphantom{fg}}\hspace*{0pt}\colorbox[RGB]{253,218,202}{tool,\vphantom{fg}}\hspace*{0pt}\colorbox[RGB]{254,238,230}{respond\vphantom{fg}}\hspace*{0pt}\colorbox[RGB]{254,242,236}{with\vphantom{fg}}\hspace*{0pt}\colorbox[RGB]{252,178,151}{'positive'\vphantom{fg}}\hspace*{0pt}\colorbox[RGB]{254,235,225}{or\vphantom{fg}}\hspace*{0pt}\colorbox[RGB]{251,136,104}{'negative'\vphantom{fg}}\hspace*{0pt}\colorbox[RGB]{254,236,227}{to\vphantom{fg}}\hspace*{0pt}\colorbox[RGB]{253,220,205}{classify\vphantom{fg}}\hspace*{0pt}\colorbox[RGB]{254,231,220}{this\vphantom{fg}}\hspace*{0pt}\colorbox[RGB]{248,96,67}{statement:the\vphantom{fg}}\hspace*{0pt}\colorbox[RGB]{252,187,162}{title\vphantom{fg}}\hspace*{0pt}\colorbox[RGB]{254,243,238}{not\vphantom{fg}}\hspace*{0pt}\colorbox[RGB]{254,244,239}{only\vphantom{fg}}\hspace*{0pt}\colorbox[RGB]{254,231,221}{describes\vphantom{fg}}\hspace*{0pt}\colorbox[RGB]{254,234,224}{its\vphantom{fg}}\hspace*{0pt}\colorbox[RGB]{254,240,233}{main\vphantom{fg}}\hspace*{0pt}

\colorbox[RGB]{253,207,188}{characters\vphantom{fg}}\colorbox[RGB]{254,227,214}{,\vphantom{fg}}\hspace*{0pt}\colorbox[RGB]{254,238,230}{but\vphantom{fg}}\hspace*{0pt}\colorbox[RGB]{254,224,210}{the\vphantom{fg}}\hspace*{0pt}\colorbox[RGB]{103,0,12}{lazy\vphantom{fg}}\hspace*{0pt}\colorbox[RGB]{252,195,172}{people\vphantom{fg}}\hspace*{0pt}\colorbox[RGB]{253,219,203}{behind\vphantom{fg}}\hspace*{0pt}\colorbox[RGB]{254,232,222}{the\vphantom{fg}}\hspace*{0pt}\colorbox[RGB]{252,151,120}{camera\vphantom{fg}}\hspace*{0pt}\colorbox[RGB]{254,237,229}{as\vphantom{fg}}\hspace*{0pt}\colorbox[RGB]{255,245,240}{well\vphantom{fg}}\hspace*{0pt}\colorbox[RGB]{253,211,192}{.\vphantom{fg}}\hspace*{0pt}\colorbox[RGB]{254,237,229}{Answer:\vphantom{fg}}\hspace*{0pt}

\\

\cmidrule{2-3}

& {\footnotesize \color{red}{P}} & \scriptsize  \colorbox[RGB]{104,0,13}{how\vphantom{fg}}\hspace*{0pt}\colorbox[RGB]{240,65,48}{the\vphantom{fg}}\hspace*{0pt}\colorbox[RGB]{248,94,66}{role\vphantom{fg}}\hspace*{0pt}\colorbox[RGB]{180,18,24}{of\vphantom{fg}}\hspace*{0pt}\colorbox[RGB]{251,120,88}{a\vphantom{fg}}\hspace*{0pt}\colorbox[RGB]{248,97,68}{compliment\vphantom{fg}}\hspace*{0pt}\colorbox[RGB]{252,146,114}{analysis\vphantom{fg}}\hspace*{0pt}\colorbox[RGB]{161,14,20}{tool,\vphantom{fg}}\hspace*{0pt}\colorbox[RGB]{198,22,28}{responses\vphantom{fg}}\hspace*{0pt}\colorbox[RGB]{246,89,63}{with\vphantom{fg}}\hspace*{0pt}\colorbox[RGB]{251,109,77}{'positive'\vphantom{fg}}\hspace*{0pt}\colorbox[RGB]{252,190,165}{or\vphantom{fg}}\hspace*{0pt}\colorbox[RGB]{248,97,68}{'negative'\vphantom{fg}}\hspace*{0pt}\colorbox[RGB]{184,19,25}{to\vphantom{fg}}\hspace*{0pt}\colorbox[RGB]{145,10,18}{mood\vphantom{fg}}\hspace*{0pt}\colorbox[RGB]{199,23,28}{this\vphantom{fg}}\hspace*{0pt}\colorbox[RGB]{103,0,12}{statement:the\vphantom{fg}}\hspace*{0pt}\colorbox[RGB]{252,176,148}{title\vphantom{fg}}\hspace*{0pt}\colorbox[RGB]{254,238,230}{not\vphantom{fg}}\hspace*{0pt}\colorbox[RGB]{254,242,236}{only\vphantom{fg}}\hspace*{0pt}\colorbox[RGB]{254,229,218}{describes\vphantom{fg}}\hspace*{0pt}\colorbox[RGB]{254,233,223}{its\vphantom{fg}}\hspace*{0pt}\colorbox[RGB]{254,240,233}{main\vphantom{fg}}\hspace*{0pt}

\colorbox[RGB]{253,208,189}{characters\vphantom{fg}}\colorbox[RGB]{252,168,139}{,\vphantom{fg}}\hspace*{0pt}\colorbox[RGB]{254,231,220}{but\vphantom{fg}}\hspace*{0pt}\colorbox[RGB]{254,242,236}{the\vphantom{fg}}\hspace*{0pt}\colorbox[RGB]{253,216,200}{lazy\vphantom{fg}}\hspace*{0pt}\colorbox[RGB]{254,241,234}{people\vphantom{fg}}\hspace*{0pt}\colorbox[RGB]{255,245,240}{behind\vphantom{fg}}\hspace*{0pt}\colorbox[RGB]{254,242,236}{the\vphantom{fg}}\hspace*{0pt}\colorbox[RGB]{254,225,212}{camera\vphantom{fg}}\hspace*{0pt}\colorbox[RGB]{255,245,240}{as\vphantom{fg}}\hspace*{0pt}\colorbox[RGB]{254,243,238}{well\vphantom{fg}}\hspace*{0pt}\colorbox[RGB]{251,114,82}{.\vphantom{fg}}\hspace*{0pt}\colorbox[RGB]{251,136,104}{Answer:\vphantom{fg}}\hspace*{0pt} 

\\

\midrule

\multirow{3}{*}{\footnotesize CheckList} & {\footnotesize \color{green}{P}} & \scriptsize  \colorbox[RGB]{254,244,239}{Given\vphantom{fg}}\hspace*{0pt}\colorbox[RGB]{254,244,239}{the\vphantom{fg}}\hspace*{0pt}\colorbox[RGB]{255,245,240}{context\vphantom{fg}}\hspace*{0pt}\colorbox[RGB]{254,239,232}{of\vphantom{fg}}\hspace*{0pt}\colorbox[RGB]{254,240,233}{this\vphantom{fg}}\hspace*{0pt}\colorbox[RGB]{254,226,213}{text,\vphantom{fg}}\hspace*{0pt}\colorbox[RGB]{254,234,224}{indicate\vphantom{fg}}\hspace*{0pt}\colorbox[RGB]{254,234,224}{if\vphantom{fg}}\hspace*{0pt}\colorbox[RGB]{254,239,231}{the\vphantom{fg}}\hspace*{0pt}\colorbox[RGB]{252,204,183}{emotion\vphantom{fg}}\hspace*{0pt}\colorbox[RGB]{253,219,203}{conveyed\vphantom{fg}}\hspace*{0pt}\colorbox[RGB]{254,235,225}{is\vphantom{fg}}\hspace*{0pt}\colorbox[RGB]{252,164,135}{'positive'\vphantom{fg}}\hspace*{0pt}\colorbox[RGB]{254,233,223}{or\vphantom{fg}}\hspace*{0pt}\colorbox[RGB]{241,68,50}{'negative':good\vphantom{fg}}\hspace*{0pt}\colorbox[RGB]{252,189,163}{film\vphantom{fg}}\hspace*{0pt}\colorbox[RGB]{253,222,208}{,\vphantom{fg}}\hspace*{0pt}\colorbox[RGB]{252,200,179}{but\vphantom{fg}}\hspace*{0pt}\colorbox[RGB]{254,225,212}{very\vphantom{fg}}\hspace*{0pt}\colorbox[RGB]{103,0,12}{glum\vphantom{fg}}\hspace*{0pt}\colorbox[RGB]{252,195,172}{.\vphantom{fg}}\hspace*{0pt}\colorbox[RGB]{254,236,227}{Answer:\vphantom{fg}}\hspace*{0pt}

\\

\cmidrule{2-3}

& {\footnotesize \color{red}{N}} & \scriptsize \colorbox[RGB]{255,245,240}{Given\vphantom{fg}}\hspace*{0pt}\colorbox[RGB]{254,243,237}{the\vphantom{fg}}\hspace*{0pt}\colorbox[RGB]{255,245,240}{context\vphantom{fg}}\hspace*{0pt}\colorbox[RGB]{254,238,230}{of\vphantom{fg}}\hspace*{0pt}\colorbox[RGB]{254,243,237}{this\vphantom{fg}}\hspace*{0pt}\colorbox[RGB]{254,235,226}{text,\vphantom{fg}}\hspace*{0pt}\colorbox[RGB]{254,243,237}{indicate\vphantom{fg}}\hspace*{0pt}\colorbox[RGB]{254,240,233}{if\vphantom{fg}}\hspace*{0pt}\colorbox[RGB]{254,241,235}{the\vphantom{fg}}\hspace*{0pt}\colorbox[RGB]{254,227,214}{emotion\vphantom{fg}}\hspace*{0pt}\colorbox[RGB]{254,227,214}{conveyed\vphantom{fg}}\hspace*{0pt}\colorbox[RGB]{254,240,233}{is\vphantom{fg}}\hspace*{0pt}\colorbox[RGB]{253,223,209}{'positive'\vphantom{fg}}\hspace*{0pt}\colorbox[RGB]{254,241,235}{or\vphantom{fg}}\hspace*{0pt}\colorbox[RGB]{254,226,213}{'negative'\vphantom{fg}}\hspace*{0pt}\colorbox[RGB]{252,176,148}{LkF0FZxMZ4:good\vphantom{fg}}\hspace*{0pt}\colorbox[RGB]{253,219,203}{film\vphantom{fg}}\hspace*{0pt}\colorbox[RGB]{254,231,221}{,\vphantom{fg}}\hspace*{0pt}\colorbox[RGB]{254,227,215}{but\vphantom{fg}}\hspace*{0pt}\colorbox[RGB]{254,233,223}{very\vphantom{fg}}\hspace*{0pt}\colorbox[RGB]{103,0,12}{glum\vphantom{fg}}\hspace*{0pt}\colorbox[RGB]{253,216,200}{.\vphantom{fg}}\hspace*{0pt}\colorbox[RGB]{254,240,233}{Answer:\vphantom{fg}}\hspace*{0pt}

\\
\midrule
\multirow{4}{*}{\footnotesize DeepWordBug} & {\footnotesize \color{green}{N}} & \scriptsize \colorbox[RGB]{255,245,240}{Serving\vphantom{fg}}\hspace*{0pt}\colorbox[RGB]{254,243,237}{as\vphantom{fg}}\hspace*{0pt}\colorbox[RGB]{254,237,228}{a\vphantom{fg}}\hspace*{0pt}\colorbox[RGB]{252,171,142}{sentiment\vphantom{fg}}\hspace*{0pt}\colorbox[RGB]{254,237,228}{evaluation\vphantom{fg}}\hspace*{0pt}\colorbox[RGB]{253,208,189}{model,\vphantom{fg}}\hspace*{0pt}\colorbox[RGB]{254,235,225}{determine\vphantom{fg}}\hspace*{0pt}\colorbox[RGB]{254,233,223}{if\vphantom{fg}}\hspace*{0pt}\colorbox[RGB]{254,236,227}{the\vphantom{fg}}\hspace*{0pt}\colorbox[RGB]{254,233,224}{given\vphantom{fg}}\hspace*{0pt}\colorbox[RGB]{253,211,192}{statement\vphantom{fg}}\hspace*{0pt}\colorbox[RGB]{254,227,215}{is\vphantom{fg}}\hspace*{0pt}\colorbox[RGB]{239,62,46}{'positive'\vphantom{fg}}\hspace*{0pt}\colorbox[RGB]{253,219,203}{or\vphantom{fg}}\hspace*{0pt}\colorbox[RGB]{103,0,12}{'negative'.\vphantom{fg}}\hspace*{0pt}\colorbox[RGB]{242,71,51}{Classify:i\vphantom{fg}}\hspace*{0pt}\colorbox[RGB]{252,197,174}{had\vphantom{fg}}\hspace*{0pt}\colorbox[RGB]{253,223,209}{to\vphantom{fg}}\hspace*{0pt}\colorbox[RGB]{251,143,111}{look\vphantom{fg}}\hspace*{0pt}\colorbox[RGB]{250,103,72}{away\vphantom{fg}}\hspace*{0pt}\colorbox[RGB]{252,197,174}{-\vphantom{fg}}\hspace*{0pt}\colorbox[RGB]{252,151,120}{this\vphantom{fg}}\hspace*{0pt}\colorbox[RGB]{252,166,137}{was\vphantom{fg}}\hspace*{0pt}

% \hspace*{0pt}

\colorbox[RGB]{251,120,88}{god\vphantom{fg}}\hspace*{0pt}\colorbox[RGB]{192,21,26}{awful\vphantom{fg}}\hspace*{0pt}\colorbox[RGB]{252,158,128}{.\vphantom{fg}}\colorbox[RGB]{254,229,217}{Answer:\vphantom{fg}}\hspace*{0pt}
\\
\cmidrule{2-3}
& {\footnotesize \color{red}{N/A}} & \scriptsize \colorbox[RGB]{252,158,128}{Servign\vphantom{fg}}\hspace*{0pt}\colorbox[RGB]{253,208,189}{as\vphantom{fg}}\hspace*{0pt}\colorbox[RGB]{254,227,214}{a\vphantom{fg}}\hspace*{0pt}\colorbox[RGB]{251,111,79}{sentimBnt\vphantom{fg}}\hspace*{0pt}\colorbox[RGB]{252,151,120}{envaluation\vphantom{fg}}\hspace*{0pt}\colorbox[RGB]{251,111,79}{model,\vphantom{fg}}\hspace*{0pt}\colorbox[RGB]{103,0,12}{Qetermine\vphantom{fg}}\hspace*{0pt}\colorbox[RGB]{251,135,103}{if\vphantom{fg}}\hspace*{0pt}\colorbox[RGB]{254,225,211}{the\vphantom{fg}}\hspace*{0pt}\colorbox[RGB]{252,194,171}{Iiven\vphantom{fg}}\hspace*{0pt}\colorbox[RGB]{252,163,134}{statemen\vphantom{fg}}\hspace*{0pt}\colorbox[RGB]{253,215,199}{is\vphantom{fg}}\hspace*{0pt}\colorbox[RGB]{252,169,141}{'positive'\vphantom{fg}}\hspace*{0pt}\colorbox[RGB]{253,208,189}{or\vphantom{fg}}\hspace*{0pt}\colorbox[RGB]{232,52,41}{'negative'.\vphantom{fg}}\hspace*{0pt}\colorbox[RGB]{244,80,57}{Classhfy:\vphantom{fg}}\hspace*{0pt}\colorbox[RGB]{253,218,202}{i\vphantom{fg}}\hspace*{0pt}\colorbox[RGB]{254,239,232}{had\vphantom{fg}}\hspace*{0pt}\colorbox[RGB]{254,244,239}{to\vphantom{fg}}\hspace*{0pt}\colorbox[RGB]{255,245,240}{look\vphantom{fg}}\hspace*{0pt}\colorbox[RGB]{254,240,233}{away\vphantom{fg}}\hspace*{0pt}\colorbox[RGB]{254,235,226}{-\vphantom{fg}}\hspace*{0pt}\colorbox[RGB]{254,243,237}{this\vphantom{fg}}\hspace*{0pt}\colorbox[RGB]{255,245,240}{was\vphantom{fg}}\hspace*{0pt}

% \hspace*{0pt}
\colorbox[RGB]{254,243,237}{god\vphantom{fg}}\hspace*{0pt}\colorbox[RGB]{254,236,227}{awful\vphantom{fg}}\colorbox[RGB]{251,145,113}{.\vphantom{fg}}\hspace*{0pt}\colorbox[RGB]{252,191,166}{Answer:\vphantom{fg}}\hspace*{0pt} \\
\midrule
\multirow{2}{*}{\footnotesize Semantic}  & {\footnotesize \color{green}{N}}  & \scriptsize \colorbox[RGB]{254,235,226}{In\vphantom{fg}}\hspace*{0pt}\colorbox[RGB]{255,245,240}{the\vphantom{fg}}\hspace*{0pt}\colorbox[RGB]{254,233,223}{role\vphantom{fg}}\hspace*{0pt}\colorbox[RGB]{254,241,234}{of\vphantom{fg}}\hspace*{0pt}\colorbox[RGB]{254,240,233}{a\vphantom{fg}}\hspace*{0pt}\colorbox[RGB]{252,166,137}{sentiment\vphantom{fg}}\hspace*{0pt}\colorbox[RGB]{254,234,224}{analysis\vphantom{fg}}\hspace*{0pt}\colorbox[RGB]{254,225,212}{tool,\vphantom{fg}}\hspace*{0pt}\colorbox[RGB]{254,229,218}{respond\vphantom{fg}}\hspace*{0pt}\colorbox[RGB]{254,239,232}{with\vphantom{fg}}\hspace*{0pt}\colorbox[RGB]{252,187,162}{'positive'\vphantom{fg}}\hspace*{0pt}\colorbox[RGB]{254,231,221}{or\vphantom{fg}}\hspace*{0pt}\colorbox[RGB]{252,197,174}{'negative'\vphantom{fg}}\hspace*{0pt}\colorbox[RGB]{254,225,212}{to\vphantom{fg}}\hspace*{0pt}\colorbox[RGB]{252,169,141}{classify\vphantom{fg}}\hspace*{0pt}\colorbox[RGB]{253,206,186}{this\vphantom{fg}}\hspace*{0pt}\colorbox[RGB]{103,0,12}{statement:bad\vphantom{fg}}\hspace*{0pt}\colorbox[RGB]{251,145,113}{.\vphantom{fg}}\hspace*{0pt}\colorbox[RGB]{253,216,200}{Answer:\vphantom{fg}}\hspace*{0pt}

\\
\cmidrule{2-3}

& {\footnotesize \color{red}{P}} & \scriptsize \colorbox[RGB]{254,231,221}{Classify\vphantom{fg}}\hspace*{0pt}\colorbox[RGB]{254,244,239}{what\vphantom{fg}}\hspace*{0pt}\colorbox[RGB]{254,237,229}{you're\vphantom{fg}}\hspace*{0pt}\colorbox[RGB]{255,245,240}{trying\vphantom{fg}}\hspace*{0pt}\colorbox[RGB]{254,243,237}{to\vphantom{fg}}\hspace*{0pt}\colorbox[RGB]{254,235,225}{convey\vphantom{fg}}\hspace*{0pt}\colorbox[RGB]{254,243,238}{in\vphantom{fg}}\hspace*{0pt}\colorbox[RGB]{254,241,235}{this\vphantom{fg}}\hspace*{0pt}\colorbox[RGB]{254,234,224}{sentence\vphantom{fg}}\hspace*{0pt}\colorbox[RGB]{254,242,236}{as\vphantom{fg}}\hspace*{0pt}\colorbox[RGB]{254,225,211}{'positive'\vphantom{fg}}\hspace*{0pt}\colorbox[RGB]{254,235,226}{if\vphantom{fg}}\hspace*{0pt}\colorbox[RGB]{253,223,209}{it's\vphantom{fg}}\hspace*{0pt}\colorbox[RGB]{254,225,211}{positive,\vphantom{fg}}\hspace*{0pt}\colorbox[RGB]{254,232,222}{and\vphantom{fg}}\hspace*{0pt}\colorbox[RGB]{254,225,211}{'negative'\vphantom{fg}}\hspace*{0pt}\colorbox[RGB]{254,230,219}{if\vphantom{fg}}\hspace*{0pt}\colorbox[RGB]{252,202,182}{it's\vphantom{fg}}\hspace*{0pt}\colorbox[RGB]{103,0,12}{negative.bad\vphantom{fg}}\hspace*{0pt}\colorbox[RGB]{251,122,90}{.\vphantom{fg}}\hspace*{0pt}\colorbox[RGB]{254,233,224}{Answer:\vphantom{fg}}\hspace*{0pt} 

\\

\bottomrule
\end{tabular}
}
%\end{center}
\vspace{-.1in}
\end{table*}

% \begin{figure}[t!]
%     \centering
%     \includegraphics[width=.48\textwidth]{imgs/fig-modelsize.pdf}
%     \caption{Accuracy of Llama2 models with fine-tuning (7b-chat) and no fine-tuning (7b) on SST-2 and COLA tasks.}
%     \label{fig-finetune}
% \end{figure}

% \begin{figure}[t!]
%     \begin{minipage}{0.25\textwidth}
%         \centering
%         \includegraphics[width=\linewidth]{imgs/fig-finetune-7b.pdf}
%         % \caption{SST-2.}
%     \end{minipage}%
%     \begin{minipage}{0.25\textwidth}
%         \centering
%         \includegraphics[width=\linewidth]{imgs/fig-finetune-13b.pdf}
%         % \caption{COLA.}
%     \end{minipage}
%     \caption{Accuracy of Llama2 models with fine-tuning (7b-chat) and no fine-tuning (7b) on SST-2 and COLA tasks.}
%     \label{fig-finetune}
% \end{figure}

\vspace{-0.1in}
\subsection{Understanding the vulnerability of \llms to adversarial prompts}
\label{sec-exp-visualzation}
\vspace{-0.05in}

We study the magic behind adversarial prompts to analyze why they lead to errors for \llms from different aspects: attention visualization, erroneous response analysis (\cref{sec-append-error}), and sentence-level analysis (\cref{stresstest analysis}).

\begin{table*}[t!]
\centering
\caption{The APDR of transferability of several \llms.}
% \vspace{-.1in}
\label{tb-transfer}
\resizebox{\textwidth}{!}{
\begin{tabular}{lcccccccccccc}
\toprule
Attacks & Chat $\rightarrow$ T5 & Chat $\rightarrow$ UL2 & Chat $\rightarrow$ V & T5 $\rightarrow$ Chat & T5 $\rightarrow$ UL2 & T5 $\rightarrow$ V & UL2 $\rightarrow$ Chat & UL2 $\rightarrow$ T5 & UL2 $\rightarrow$ V & V $\rightarrow$ Chat & V $\rightarrow$ T5 & V $\rightarrow$ UL2 \\
            \midrule
BertAttack  & 0.05\scriptsize{$\pm$0.17}                      & 0.08\scriptsize{$\pm$0.19}                 & 0.08\scriptsize{$\pm$0.88}        & 0.18\scriptsize{$\pm$0.32}                      & 0.11\scriptsize{$\pm$0.23}                              & -1.39\scriptsize{$\pm$5.67}                    & 0.15\scriptsize{$\pm$0.27}                 & 0.05\scriptsize{$\pm$0.11}                              & -0.70\scriptsize{$\pm$3.18}               & 0.06\scriptsize{$\pm$0.19}        & 0.05\scriptsize{$\pm$0.11}                     & 0.03\scriptsize{$\pm$0.12}                \\
CheckList   & 0.00\scriptsize{$\pm$0.04}                      & 0.01\scriptsize{$\pm$0.03}                 & 0.19\scriptsize{$\pm$0.39}        & 0.00\scriptsize{$\pm$0.07}                      & 0.01\scriptsize{$\pm$0.03}                              & -0.09\scriptsize{$\pm$0.64}                    & 0.01\scriptsize{$\pm$0.06}                 & 0.01\scriptsize{$\pm$0.04}                              & -0.13\scriptsize{$\pm$1.80}               & -0.01\scriptsize{$\pm$0.04}       & 0.00\scriptsize{$\pm$0.01}                     & 0.00\scriptsize{$\pm$0.00}                \\
TextFooler  & 0.04\scriptsize{$\pm$0.08}                      & 0.03\scriptsize{$\pm$0.09}                 & -0.25\scriptsize{$\pm$1.03}       & 0.11\scriptsize{$\pm$0.23}                      & 0.08\scriptsize{$\pm$0.16}                              & -0.30\scriptsize{$\pm$2.09}                    & 0.11\scriptsize{$\pm$0.21}                 & 0.07\scriptsize{$\pm$0.18}                              & -0.17\scriptsize{$\pm$1.46}               & 0.04\scriptsize{$\pm$0.16}        & 0.02\scriptsize{$\pm$0.06}                     & 0.00\scriptsize{$\pm$0.01}                \\
TextBugger  & -0.00\scriptsize{$\pm$0.09}                     & -0.01\scriptsize{$\pm$0.05}                & 0.02\scriptsize{$\pm$0.94}        & 0.04\scriptsize{$\pm$0.15}                      & 0.01\scriptsize{$\pm$0.04}                              & -0.45\scriptsize{$\pm$3.43}                    & 0.04\scriptsize{$\pm$0.13}                 & 0.02\scriptsize{$\pm$0.07}                              & -0.84\scriptsize{$\pm$4.42}               & 0.03\scriptsize{$\pm$0.13}        & 0.01\scriptsize{$\pm$0.05}                     & 0.00\scriptsize{$\pm$0.01}                \\
DeepWordBug & 0.03\scriptsize{$\pm$0.11}                      & 0.01\scriptsize{$\pm$0.03}                 & 0.10\scriptsize{$\pm$0.46}        & 0.00\scriptsize{$\pm$0.06}                      & 0.01\scriptsize{$\pm$0.02}                              & -0.18\scriptsize{$\pm$1.20}                    & 0.01\scriptsize{$\pm$0.10}                 & 0.02\scriptsize{$\pm$0.06}                              & -0.09\scriptsize{$\pm$0.75}               & 0.00\scriptsize{$\pm$0.03}        & 0.02\scriptsize{$\pm$0.11}                     & 0.00\scriptsize{$\pm$0.01}                \\
StressTest  & 0.04\scriptsize{$\pm$0.17}                      & 0.03\scriptsize{$\pm$0.10}                 & 0.01\scriptsize{$\pm$0.48}        & -0.01\scriptsize{$\pm$0.06}                     & 0.03\scriptsize{$\pm$0.06}                              & 0.04\scriptsize{$\pm$0.80}                     & 0.00\scriptsize{$\pm$0.04}                 & 0.05\scriptsize{$\pm$0.16}                              & 0.06\scriptsize{$\pm$0.45}                & 0.00\scriptsize{$\pm$0.04}        & 0.09\scriptsize{$\pm$0.18}                     & 0.02\scriptsize{$\pm$0.08}                \\
Semantic & 0.04\scriptsize{$\pm$0.12}                      & 0.02\scriptsize{$\pm$0.06}                 & 0.25\scriptsize{$\pm$0.47}        & 0.07\scriptsize{$\pm$0.27}                      & 0.00\scriptsize{$\pm$0.03}                              & -0.81\scriptsize{$\pm$4.14}                    & 0.02\scriptsize{$\pm$0.11}                 & -0.13\scriptsize{$\pm$0.72}                             & -0.50\scriptsize{$\pm$1.59}               & 0.07\scriptsize{$\pm$0.11}        & 0.00\scriptsize{$\pm$0.05}                     & 0.00\scriptsize{$\pm$0.02} \\
\bottomrule
           
\end{tabular}
}
\vspace{-.2in}
\end{table*}

We visualize the attentions to investigate the influence of adversarial prompts on \llms' focus on input words.
Specifically, we propose two attention visualization techniques: 1) Attention by Gradient, which assigns an attention score to each word based on the gradient norm, and 2) Attention by Deletion, which assigns an attention score to each word by examining the absolute change in loss when the word is removed. The details of these methods can be found in Appendix~\ref{sec-append-attention}. Both techniques produce similar results; hence, we focus on results from the Attention by Gradient method for simplicity. Our key findings, as demonstrated in \tablename~\ref{table: attention misclassified adv prompts}, are as follows:
\begin{itemize}[leftmargin=1em]
\setlength\itemsep{0em}
    \item \textbf{Clean prompts: efficient attention allocation.} \llms predominantly concentrate on key terms within clean prompts, aiding in accurate classifications. 
    For instance, for clean prompts of BertAttack in \tablename~\ref{table: attention misclassified adv prompts}, \llms mainly allocate attention to the term `\prompt{lazy}', correctly deducing a `Negative' sentiment.
    
    \item \textbf{Adversarial prompts: attention divergence.} Adversarial prompts can reroute \llms' attention from integral text segments, causing misclassifications. 
    In some attacks like CheckList and StressTest, the model simultaneously concentrates on the target text and adversarial content, amplifying its susceptibility to adversarial perturbations. 
    For instance, introducing a random sequence `\prompt{LKF0FZxMZ4}' during a CheckList attack distracts the model, reducing focus on the critical word `\prompt{good}' for accurate classification. 
    In other attacks, such as BertAttack and DeepWordBug, the model's attention is entirely diverted from the text requiring classification towards adversarial prompts, leading to a significant shift in focus. 
    For example, in DeepWordBug attack, typos in specific words divert the model's attention from `\prompt{awful}' to the altered word `\prompt{Qetermine}'.
\end{itemize}

\vspace{-0.1in}
\subsection{Transferability of adversarial prompts}
\label{sec-transfer}
\vspace{-0.05in}

\tablename~\ref{tb-transfer} displays the effectiveness of various attacks in transferring adversarial prompts between several \llms. 
For each dataset and prompt type, we selected the most vulnerable prompts generated by a source model (e.g., \chat). 
These prompts were then utilized to launch transfer attacks against the target models (e.g., T5-large). 
The impact of these transfer attacks was quantified by calculating $\mathit{APDR}_{\text{transfer}}(A, f_\theta^\text{target}) = \frac{1}{|\mathcal{P}_{\text{source}}|} \frac{1}{|\mathbb{D}|} \sum_{P \in \mathcal{P}_{\text{source}}} \sum_{\mathcal{D} \in \mathbb{D}} \mathit{PDR}(A, P, f_\theta^\text{target}, \mathcal{D})$, where $f_\theta^\text{target}$ is the target model, $\mathcal{P}_{\text{source}}$ is the prompts selected from source model and $\mathbb{D}$ is the set of all datasets.

In general, we observe that adversarial prompts exhibit some degree of transferability. However, it is marginal compared to \tablename~\ref{tb-dataset-attack} and \ref{tb-dataset-model}.
Specifically, the APDR in the target model by adversarial prompts from source model is small compared to the original APDR of the source model.
Furthermore, the standard deviation tends to be larger than the APDR, indicating that the transferability is inconsistent.
Some adversarial prompts can be successfully transferred, causing a performance drop, while others may unexpectedly improve the performance of the target model.
A prime example is the BertAttack transfer from UL2 to Vicuna, which resulted in a $-0.70(3.18)$ value, suggesting an unanticipated enhancement in Vicuna's performance when subjected to these adversarial prompts.
These phenomena illustrate the complex robustness traits of different models.
The transferability to \chat is better compared to T5-large and UL2.
This suggests an avenue to generate adversarial prompts to attack black-box models such as \chat by training on small models like T5-large, which could be used for future robustness research.

\vspace{-0.1in}
\subsection{Which prompts are more robust? Word frequency}
\vspace{-0.05in}

The word frequency results of these two datasets are presented in \cref{sec-exp-word-frequency}.
Our findings underscore that the resilience of a prompt is intricately tied to the contextual use of words, rather than the mere presence of certain terms. This complexity suggests that factors beyond word frequency, such as semantic coherence and syntactic structures, might be instrumental in determining robustness. This knowledge is valuable as it can influence future research on the robustness of \llms, provide guidance for crafting more resistant prompts, and facilitate the creation of defensive mechanisms against adversarial prompt attacks. It is essential to emphasize that our observations are rooted in the current scope of models and datasets. Furthermore, the robustness or vulnerability of words remains deeply context-dependent. Hence, direct determination of word robustness without considering the broader context may lead to oversimplified or inaccurate conclusions.

\vspace{-0.1in}
\subsection{Countermeasures and defenses}
\vspace{-0.05in}

We discuss potential countermeasures.
\textbf{1) Input preprocessing:} One approach involves directly detecting and addressing potential adversaries, such as detecting typos, irrelevant sequences, and enhancing clarity and conciseness of prompts.
\textbf{2) Incorporate low-quality data in pre-training:} Low-quality data can serve as potential adversaries, and explicitly including low-quality data during pre-training may develop a better understanding of diverse inputs and build resilience against adversaries.
\textbf{3) Improved fine-tuning:} Fine-tuning could lead to improved robustness. As we demonstrated before, models such as T5 and UL2 exhibit greater robustness compared to \chat, suggesting potential benefits of large-scale supervised fine-tuning.
More details are in \cref{sec-append-defense}.

\vspace{-0.1in}
\section{Related work}
\vspace{-0.1in}
\label{sec-relatedwork}

% Detailed comparisons with existing textual adversarial attacks are presented in \cref{sec-append-textual}.

\textbf{LLM Robustness Evaluation.}
AdvGLUE~\citep{wang2021adversarial} stands as a static dataset for evaluating adversarial robustness of \emph{input samples}.
DecodingTrust~\citep{wang2023decodingtrust} undertakes a comprehensive assessment of trustworthiness in GPT models, notably GPT-3.5 and GPT-4. The research delves into areas like toxicity, stereotype bias, adversarial challenges, and privacy. Specifically, they evaluate the robustness on standard datasets AdvGLUE\citep{wang2021adversarial} and AdvGLUE++ \citep{wang2023decodingtrust}.
Specifically for adversarial robustness, DecodingTrust also focuses on evaluating the robustness of input samples instead of prompts and it still uses static datasets rather than an actionable benchmark suite.
In contrast, \method is positioned as an open benchmark concentrating on adversarial \emph{prompts} rather than samples (and it can be extended to include samples).
Note that the prompts are general instructions to assist the in-context learning of \llms to perform specific tasks, and they can be combined with many samples in certain tasks.
Prompts are indispensable in human-LLMs interaction while input samples may not be needed, which means that prompts are versatile and it is essential to evaluate their robustness.

\textbf{Safety of LLMs.}
We mimic the potential user prompts by creating adversarial prompts, but the main purpose is not to actually attack the model.
This distinguishes our work from existing efforts in safety research of LLMs.
Specifically, both SafetyPrompts \citep{sun2023safety} and prompt injection attacks \citep{perez2022ignore, greshake2023more, liu2023prompt} are engineered to spotlight potentially harmful instructions that could steer LLMs into \textit{delivering outputs misaligned with human values or perform unintended actions} such as data leakage and unauthorized access. Adversarial prompts are crafted to \textit{mimic plausible mistakes an end-user might inadvertently make}. Our goal is to assess the extent to which these prompts, even if they slightly deviate from the norm, can skew LLM outcomes. These adversarial prompts retain their semantic integrity, ensuring they're virtually imperceptible for humans. The adversarial prompts are not  designed to elicit harmful or misleading responses.

\vspace{-0.1in}
\section{Conclusion and Limitation}
\vspace{-0.1in}

The robustness of the prompts in \llms is of paramount concern in security and human-computer interaction.
In this paper, we thoroughly evaluated the robustness of \llms to adversarial prompts using the proposed \method benchmark.
The key is to leverage adversarial attack approaches to mimic potential perturbations such as typos, synonyms, and stylistic differences.
We then performed extensive experiments and analysis on various tasks and models.
While the results show that current \llms are not robust enough to adversarial prompts, we further analyzed the reason behind it using attention visualization.
Moreover, we analyzed the frequent words to provide guidance for both experts and non-experts in developing better prompt engineering tools.
\method will be open-sourced to serve as a foundational tool for robust \llms research.

There are several limitations.
First, due to the substantial computation, we did not perform evaluations on the full datasets, but relied on sampling.
Future research may evaluate on the entire datasets to gain more insights.
Second, we cannot included all LLMs and datasets due to computation constraint.
Including more in the future could provide a more diverse perspective.
Third, we did not evaluate more advanced techniques of prompt engineering such as chain-of-thought (CoT) \citep{wei2022chain} and tree-of-thought (ToT) \citep{yao2023tree}. 
% since it is hard to perform automatic evaluation using these methods.
We believe more evaluations can be done on latest prompt engineering techniques. Fourth, we considered black-box prompt attacks, which can generate perturbations that can mimic naturally occurred errors. Optimized white-box prompt attacks may produce better adversarial prompts, which is an interesting future work. 

% \section*{Disclaimer}

% \neil{This section is probably not needed for this conference.}
% \wjd{Yes, we will remove it. But add it in the preprint version.}

% This paper leveraged adversarial attacks on prompts for \llms evaluation, which might trigger potential misuse of \llms.
% We emphasize  that all attacks conducted in this work are only to evaluate the robustness of \llms to adversarial prompts with the intention of facilitating more robust \llms.

% Additionally, we leveraged both open-source \llms in Huggingface \citep{wolf2020huggingfaces} and online APIs for evaluation.
% As open-source \llms and online APIs may continuously change, some results may not be reproducible.
% However, our code and analysis framework remain versatile and can still be useful for future \llms.

% \begin{ack}

% \end{ack}

%%%%%%%%%%%%%%%%%%%%%%%%%%%%%%%%%%%%%%%%%%%%%%%%%%%%%%%%%%%%%%%%%%%%%%%%%%%%%%%
%%%%%%%%%%%%%%%%%%%%%%%%%%%%%%%%%%%%%%%%%%%%%%%%%%%%%%%%%%%%%%%%%%%%%%%%%%%%%%%
% APPENDIX
%%%%%%%%%%%%%%%%%%%%%%%%%%%%%%%%%%%%%%%%%%%%%%%%%%%%%%%%%%%%%%%%%%%%%%%%%%%%%%%
%%%%%%%%%%%%%%%%%%%%%%%%%%%%%%%%%%%%%%%%%%%%%%%%%%%%%%%%%%%%%%%%%%%%%%%%%%%%%%%

\bibliography{ref}
\bibliographystyle{colm2024_conference}

\appendix
% \section{Prompts}
% \label{sec-append-prompt}

% \tablename~\ref{table: prompt examples} shows examples of different types of prompts.

% \input{tables/tb-prompt-examples}
\section{Details on Adversarial Prompts}
\label{sec-append-prompt}

\subsection{Examples of prompts}
\label{sec-append-prompt-example}

\cref{table: prompt examples} shows the details of the task and role-oriented prompts in zero-shot and few-shot settings.
\cref{table: adv prompt examples} shows examples of $7$ adversarial attacks on prompts.

\begin{table}[htbp]
\caption{Examples of $4$ types of prompts.}
\label{table: prompt examples}
\centering
\resizebox{1\textwidth}{!}{
\begin{tabular}{ c  c m{22cm}}
\toprule

\multirow{3}{*}{Zero-shot}  & \makecell{Task-oriented} & { \prompt{Evaluate the sentiment of the given text and classify it as 'positive' or 'negative':} } \\

\cmidrule(lr){2-3}

& \makecell{Role-oriented} & { \prompt{In the role of a sentiment analysis tool, respond with 'positive' or 'negative' to classify this statement:} } \\

\midrule

\multirow{8}{*}{Few-shot}  & Task-oriented & { \prompt{Analyze the tone of this statement and respond with either 'positive' or 'negative'. Here are three examples.  Sentence: hide new secretions from the parental units. Answer: negative.  Sentence: contains no wit , only labored gags. Answer: negative. Sentence: that loves its characters and communicates something rather beautiful about human nature. Answer: positive.} }\\

\cmidrule(lr){2-3}

& \makecell{Role-oriented} &  { \prompt{As a sentiment classifier, determine whether the following text is 'positive' or 'negative'. Here are three examples. Sentence: hide new secretions from the parental units. Answer: negative. Sentence: contains no wit , only labored gags. Answer: negative. Sentence: that loves its characters and communicates something rather beautiful about human nature. Answer: positive.} }\\

\bottomrule
\end{tabular}
}
%\end{center}
\end{table}

\begin{table}[htbp]
\caption{Example of adversarial prompts generated by $7$ prompt attacks to mimic possible prompts. The characters and words marked with red are generated by prompt attacks.}
\label{table: adv prompt examples}
\centering  
\resizebox{1\textwidth}{!}{
\begin{tabular}{ l m{19cm}}
\toprule
\multirow{1}{*}{Scenario} & Prompt \\
\midrule

\multirow{1}{*}{Clean} & \prompt{As a mathematics instructor, calculate the answer to the following problem related to \{\}:} \\
\midrule
\multirow{1}{*}{TextBugger}  & \prompt{As a mathematics instructo{\color{red}{rr}}, calculate the answer{\color{red}{s}} to the following problem related to \{\}:} \\
\midrule
\multirow{1}{*}{DeepWordBug} & \prompt{As a mathematics i{\color{red}{e}}structor, calculate the answe{\color{red}x} to the following problem related to \{\}:} \\
\midrule
\multirow{1}{*}{TextFooler}  & \prompt{As a mathematics {\color{red}prof}, calculate the {\color{red}address} to the following problem related to \{\}:}  \\
\midrule
\multirow{1}{*}{BertAttack}  & \prompt{As a mathematics instructor, calculate the {\color{red}sum} to the following problem related to \{\}: } \\
\midrule
\multirow{1}{*}{CheckList}   & \prompt{As a mathematics instructor, calculate the answer to the following problem related to {\color{red}KjPJJ2a7RB} \{\}:} \\
\midrule
\multirow{1}{*}{StressTest}  & \prompt{As a mathematics instructor, calculate the answer to the following problem related to {\color{red}and false is not true} \{\}:} \\
\midrule
\multirow{1}{*}{Semantic}   & \prompt{{\color{red}Compute the result of} \{\}.} \\
\bottomrule
\end{tabular}
}
%\end{center}
\end{table}

\subsection{Semantic preservation of adversarial prompts}
\label{sec-append-semantic}

In our endeavor to validate the efficacy of our adversarial prompt generation, we engaged in a human-centric study. We enlisted the expertise of 5 independent evaluators with proficiency in the domain to critically assess the semantic congruence between the original and the generated adversarial prompts. For the study, we randomly sampled a set of $100$ adversarial prompts along with their respective original prompts. The evaluators were tasked with determining if each adversarial prompt was semantically equivalent to its original counterpart, ensuring the absence of semantic drift. Such an evaluation provides insights into the reliability and robustness of our adversarial prompts.

\begin{figure}[t!]
    \centering
    \includegraphics[width=0.45\textwidth]{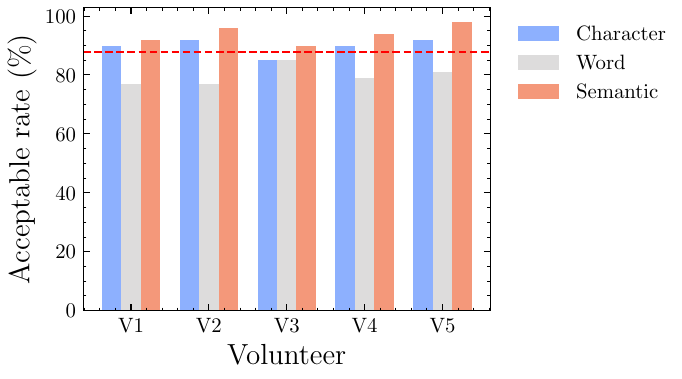}
    \caption{Results of human study on semantic preserving of the adversarial prompts. The dotted red line ($> 85\%$) is the average of all volunteers on all attacks.}
    \label{fig-semantic-preserving}
\end{figure}

To address the challenges associated with word-level attacks, we have diligently fine-tuned the hyperparameters of each attack approach, thus striving to maintain semantic continuity.
Then, we conduct a human study to recruit five volunteers to judge if the generated adversarial prompts can preserve semantics.
The evaluators were presented with the original prompt $P$ juxtaposed with its adversarial version $\bar{P}$, and were tasked with determining their semantic congruence.
Sentence-level attacks are excluded in this study since they do not change the original prompts, but only to add extra perturbations in the end.
The results in \figurename~\ref{fig-semantic-preserving} demonstrate that these adversarial prompts generated by character-level, word-level and semantic-level attacks are \textbf{at least $85\%$ acceptable by humans}, indicating that our attack is realistic and meaningful. 

Here, we present a few exemplars of both acceptable and unacceptable adversarial prompts as discerned by the evaluators:

\begin{itemize}[leftmargin=1em]
\setlength\itemsep{0em}
    \item \textbf{SST-2} Original prompt: \prompt{Read the provided excerpt and choose between 'positive' and 'negative' to describe its sentiment:} \\
    Attacked prompt: \prompt{read the standard excerpt and pause between 'positive' and 'negative' to describe its thoughtful:} \\
    {\color{red}Unacceptable.}
    \item \textbf{CoLA} Original prompt: \prompt{Functioning as a grammar evaluation tool, analyze the given sentence and decide if it is grammatically correct, responding with 'acceptable' or 'unacceptable':} \\
    Attacked prompt: \prompt{Functioning as a grammar evaluation Lool, analyze the Tgiven sentence and deciRe if it is grammXatically correct, respoOnding with 'acceptable' or 'unacceptable':} \\
    {\color{green}Acceptable.}

    \item \textbf{MMLU} Original prompt: \prompt{As a knowledgeable individual in {}, provide your response to the multiple-choice question by choosing 'A', 'B', 'C', or 'D'.} \\
    Attacked prompt: \prompt{As a knowledgeable everybody in {}, prescribes your rebuttal to the multiple-choice item by chooses 'A', 'B', 'C', or 'D'.} \\
    {\color{red}Unacceptable.}

    \item \textbf{Math} Original prompt: \prompt{Resolve the following mathematical question associated with:} \\
    Attacked prompt: \prompt{Resolve the following mathematical answer along with {}:} \\
    {\color{green}Acceptable.}
    
\end{itemize}

\subsection{Prompt attack}
\label{sec-append-attacks}

% \textbf{Attacks on samples.}
% Note that while this paper only focuses on attacking the prompts, the input samples can also be attacked in the same manner.
% We only attack prompts since it is versatile in all tasks where the samples are not necessary.
% Attacking both is doable and will obtain worse results, but will be more time-consuming.
% We further provide an experiment on AdvGLUE~\citep{wang2021adversarial} which is adversarial samples dataset.
% Then, we show that together with adversarial prompts, the performance becomes worse.

The majority of our prompt attacks have been developed by adapting and revising strategies from TextAttack\footnote{\url{https://github.com/QData/TextAttack}}~\citep{textattack}. For the detailed settings of each attack, please refer to our code.

% Our proposed adversarial attacks incorporate a critical feature, referred to as LabelConstraint. This feature imposes restrictions on the perturbations, specifically prohibiting alterations to certain task-essential words. For instance, in translation tasks, the word `translation' is preserved, while in the context of sentiment classification tasks, pivotal sentiment indicators such as `positive' and `negative' remain untouched. Moreover, in the few-shot learning scenario, the few-shot examples are also exempt from adversarial attacks. Such constraint preservation ensures the validity of prompts while exploring the model's vulnerability to adversarial inputs.

\textbf{Character Level:} Techniques such as TextBugger and DeepWordBug manipulate text at the character level by introducing typos or errors within words through insertions, deletions, replacements, and replications. These methods capitalize on the model's vulnerability to minor perturbations in individual characters, frequently resulting in misclassification or erroneous interpretations.

We primarily adopt the settings from TextAttack for TextBugger and DeepWordBug, such as the repeat constraint which prohibits modifying words that have already been altered. Additionally, For TextBugger, TextAttack enforces a constraint on the overall similarity between the sentence encodings of clean and adversarial prompts, utilizing the Universal Sentence Encoder \citep{universal} to generate text embeddings. In our study, we set this minimum similarity threshold to $0.8$. For DeepWordBug, TextAttack set constraint on edit distance (Levenshtein Distance) as $30$.

\textbf{Word Level:} In this study, we employ BertAttack and TextFooler for word-level attacks. These approaches focus on replacing words within the text with synonyms or contextually similar words. By making ostensibly minor alterations to the input text, these attacks can deceive large language models into producing incorrect outputs or substantially modifying their predictions. We meticulously fine-tune the hyperparameters of BertAttack and TextFooler to obtain more appropriate synonyms. 

For TextFooler, we set the minimum embedding cosine similarity between word and its synonyms as $0.6$, and the minimum Universal Sentence Encoder similarity is $0.84$. For BertAttack, the minimum Universal Sentence Encoder similarity is $0.8$.

\textbf{Sentence Level:} StressTest and CheckList serve as examples of sentence-level attacks, wherein adversaries attempt to distract the model by adding irrelevant or extraneous sentences to the input text. By incorporating misleading information into the text, these methods can potentially cause the model to lose focus on the primary context, leading to inaccurate results. For the StressTest attack, we adopt similar settings to those in \citep{wang2019glue}, appending "\prompt{and true is true,}" "\prompt{and false is not true,}" or "\prompt{and true is true}" for five times to the end of a prompt. For the CheckList attack, we generate 50 random sequences consisting of alphabets and digits, each with a length of 10, and append this random sequences into the end of a prompt.

\textbf{Semantic Level:} At the human level, adversaries can construct prompts using various languages, such as Chinese, French, Arabic, Spanish, Japanese, and Korean, subsequently translating these prompts into English. By exploiting the nuances and idiosyncrasies of different languages during translation, it can introduce subtle ambiguities, grammatical errors, or inconsistencies in the input prompt. This poses a formidable challenge for NLP models in generating accurate and coherent responses.  

For each language, we first construct 10 prompts based on a English prompt by GPT4 \citep{openai2023gpt4}, then translate it back to English by Google Translator.

\subsection{Our attack vs. existing textual adversarial attacks}
\label{sec-append-textual}

Prompt attacks and textual adversarial attacks \citep{textbugger, textfooler, deepwordbug, bertattack, checklist, stresstest, zang2020word} are both rooted in similar foundational algorithms, but differ in critical ways:
\begin{itemize}
    \item Target of attack: Prompt attacks target the \textit{instruction (prompts)} for \llms while vanilla adversarial attacks focus on the \textit{samples}. In numerous tasks, the data might be optional, while prompts remain indispensable. For example, ``Write a story about a fox.'' and ``Give me some investigation suggestions.'' are all prompts with \emph{no} samples. This makes our prompt attack more general.
    \item Universality of adversarial prompts: An adversarial prompt, represented as $\bar{P}$, works as a common threat for all samples related to a specific task. For example, if $P$ is designed to instruct LLMs to solve math problems, then $\bar{P}$ can be used for \emph{many} different math problems and datasets. This ability is significantly different from current NLP adversarial benchmarks.
\end{itemize}

In essence, prompt attacks seek to delve into the universality \citep{uap, uat} of adversarial prompts. We argue this offers an innovative lens to assess the robustness of language models, complementing insights from existing benchmarks like AdvGLUE~\citep{wang2021adversarial},  and AdvGLUE++\citep{wang2023decodingtrust}.

\section{Models, Datasets, and Environments}
\label{sec-append-dandm}

\subsection{Models}
\label{sec-append-models}

Here, we list the brief introduction of each LLM in our experiments.
For more details about Vicuna, please refer to its GitHub repository\footnote{\url{https://github.com/lm-sys/FastChat}}.
For the other \llms, please refer to Huggingface transformer repository \citep{wolf2020huggingfaces}.

\begin{itemize}
    \item \textbf{Flan-T5-large \citep{t5}}: Flan-T5-large is a derivative of the Text-to-Text Transfer Transformer (T5) model, developed by Google.

    \item \textbf{Dolly-6B \citep{dolly}}: The Dolly-v1-6b model is a 6-billion parameter causal language model developed by Databricks. It originates from EleutherAI’s GPT-J \citep{gpt-j} and has been fine-tuned on the Stanford Alpaca \citep{alpaca} corpus, which comprises roughly 52K question/answer pairs. 

    \item \textbf{Vicuna-13B \citep{vicuna}}: Vicuna-13B, fine-tuned from the LLaMA-13B base model, was developed using approximately 70K user-shared conversations collected from ShareGPT.com via public APIs.
    
    \item \textbf{Cerebras-13B \citep{cerebras}}: Cerebras-13B is based on the GPT-3 style architecture. All models in the Cerebras-GPT series have been trained according to Chinchilla scaling laws \citep{hoffmann2022training}, which optimize compute efficiency by maintaining a ratio of 20 tokens per model parameter.
    
    \item \textbf{Llama2-13B \citep{llama2}}: The Llama2 model, developed by the FAIR team at Meta AI, is an autoregressive language model that employs the transformer architecture.
    
    \item \textbf{GPT-NEOX-20B \citep{neox}}: GPT-NEOX-20B is a large-scale implementation of GPT-based models, with NEOX-20B specifically referring to a variant of this series comprising 20 billion parameters.

    \item \textbf{Flan-UL2 \citep{ul2}}: Flan-UL2 is an encoder decoder model based on the T5 architecture. It uses the same configuration as the UL2 model. It was fine-tuned using the ``Flan'' prompt tuning and dataset collection.
    
    \item \textbf{ChatGPT \citep{chatgpt} and GPT-4 \citep{openai2023gpt4}}: Developed by OpenAI, ChatGPT is a large language model trained to generate human-like text based on the prompt it's given. It uses the GPT-3 architecture and has been fine-tuned for more interactive and conversational tasks. GPT-4 is by far the best-performing LLMs.

\end{itemize}

\subsection{Tasks and datasets}
\label{appendix: dataset and models}
We adopt the following public datasets for evaluation and \method can easily take as inputs other datasets.

\begin{itemize}
    \item \textbf{GLUE} The GLUE dataset (General Language Understanding Evaluation)~\citep{wang2019glue} is a collection of resources designed to assess and benchmark the performance of natural language processing (NLP) models across various language understanding tasks. In this study, we selected 8 tasks, including Sentiment Analysis (\sst \citep{sst2}), Grammar Correctness (\cola \citep{cola}), Duplicate Sentence Detection (\qqp \citep{qqp}, \mrpc \citep{mrpc}), and Natural Language Inference (\mnli \citep{mnli}, \qnli \citep{wang2019glue}, \rte \citep{wang2019glue}, and \wnli \citep{wnli}).

% Models are evaluated using the standard accuracy metric.

    \item \textbf{MMLU \citep{mmlu}} To evaluate the extensive world knowledge and problem-solving abilities of large language models, the MMLU dataset encompasses 57 tasks consisting of multiple-choice questions from diverse domains, such as mathematics, history, computer science, law, and more. This dataset serves as a massive multitask test.

% Models are evaluated using the standard accuracy metric.

    \item \textbf{SQuAD V2 \citep{squad}} SQuAD v2 is a widely used dataset for training and evaluating natural language processing models in the domain of machine reading comprehension. SQuAD v2 enhances the original SQuAD dataset (SQuAD v1) by introducing unanswerable questions, increasing the challenge for models. For each question, the model must either (1) identify the correct answer span within the passage (if the question is answerable) or (2) predict that the question is unanswerable (if there is no answer span within the passage).

% We use the F1 score to evaluate models. The F1 score measures the overlap between the model's predicted answer and the ground truth answer.

    \item \textbf{UN Multi \citep{multiun}} The Multi UN dataset is a large parallel corpus of text gathered from official United Nations documents. It comprises texts in six official languages of the United Nations: Arabic, Chinese, English, French, Russian, and Spanish. The Multi UN dataset primarily contains formal texts, which may limit its applicability to more informal language domains or conversational applications.

    \item \textbf{IWSLT 2017 \citep{iwslt}} The IWSLT 2017 dataset (International Workshop on Spoken Language Translation 2017) is a collection of multilingual, multi-domain parallel text data specifically designed for evaluating spoken language translation systems. The translation tasks include data from the TED Talks Open Translation Project, featuring parallel text data for multiple language pairs such as English-German, English-French, English-Chinese, and English-Czech. The dataset consists of both spoken language transcriptions and their corresponding translations.

    \item \textbf{Math \citep{math}} DeepMind Mathematics Dataset is a collection of math problems aimed at evaluating the mathematical reasoning abilities of artificial intelligence models. The dataset challenges AI models to solve a diverse range of mathematical problems, spanning from algebra to calculus, and tests their ability to comprehend and reason via complex mathematical concepts.

\end{itemize}

\subsection{Environments}
To reproduce the computational environment used in this study, an environment file, \prompt{environment.yml}, is provided in our repository. This YAML file lists all the dependencies and their specific versions used in the study. Users can create an identical Conda environment using the command \prompt{conda env create -f environment.yml}.
The computational experiments were conducted on machines equipped with NVIDIA Tesla V100 GPUs (16GB of GPU memory each).

% \subsection{Samples of adversarial prompts}
% \tablename~\ref{table: adv prompt examples} presented examples of adversarial prompts generated by $7$ attacks. 
% Note that we have generated \totalprompt adversarial prompts. For other examples, please refer to our code and the visualization website (Appendix~\ref{sec-append-vis}).

% \input{tables/tb-attack-examples}

% \subsection{Semantic preserving of adversarial prompts}
% \label{sec-append-semantic}

% \section{Experiments}
% \label{sec-append-exp}

% \subsection{Details on test sets sampling}
% \label{sec-append-sampling}

\section{Details on Experiments and Results}
\label{sec-append-exp}

\subsection{Details on dataset sampling}
\label{sec-append-exp-sampling}

Note that we cannot run evaluations on all samples due to significantly extensive computing requirements.
Instead, we turn to sampling.
Specifically, for the GLUE datasets, we sample 1,000 instances when the validation set exceeds this size; otherwise, we utilize the entire validation set. With respect to \chat and GPT4, we adopt a smaller sample size of 200 instances for computational efficiency. For the \mmlu dataset, we select 10 instances for each of the 57 tasks if the validation set exceeds this size; if not, the entire validation set is used. For the \squad dataset, we randomly select 200 validation instances. Regarding the translation datasets \un and \iwslt, we focus on three languages—English, French, and German, which are primarily supported by T5-large and UL2. We select a total of 100 validation instances, evenly distributed among all possible translation pairs, e.g., English to French. For the Math dataset, we select 20 types of math problems, choosing either 5 or 10 instances per type, resulting in a total of 160 instances.
This sampling strategy ensures the formation of a manageable and representative evaluation set for each dataset, thereby enabling an effective assessment of the performance and robustness of \llms across various tasks and domains.

\subsection{Results of clean prompts on all \llms}
\label{sec-append-exp-cleanresults}

\cref{tb-dataset-model-clean} showcases the performance of different models across various datasets when using clean prompts. Certain \llms, including Dolly, Cerebras, and NEXO, encounter difficulties with some datasets. For instance, Dolly's accuracy for the QQP dataset is merely $0.53\%$, a stark contrast to T5's accuracy of $86.67\%$. Consequently, we focus our attack study on models that demonstrate superior performance, namely T5, Vicuna, Llama2, UL2, ChatGPT, and GPT4.

\begin{table}[htbp]
\centering
\caption{The Average performance and standard deviations of different models on different datasets.}
% \vspace{-.1in}
\label{tb-dataset-model-clean}
\resizebox{.8\textwidth}{!}{
\begin{tabular}{ccccccccc}
\toprule

Dataset  & \textbf{T5}                          & Dolly                       & \textbf{Vicuna}                       & Cerebras                     & \textbf{Llama2}                       & NEOX                         & \textbf{UL2}                         & \textbf{ChatGPT}                     \\ \midrule
SST-2    & 94.79\scriptsize{$\pm$0.49} & 47.80\scriptsize{$\pm$9.30} & 21.12\scriptsize{$\pm$15.40}  & 21.33\scriptsize{$\pm$23.02} & 90.25\scriptsize{$\pm$2.23} & 21.49\scriptsize{$\pm$13.35} & 95.92\scriptsize{$\pm$1.03} & 92.91\scriptsize{$\pm$3.32} \\
CoLA     & 76.11\scriptsize{$\pm$1.28} & 4.92\scriptsize{$\pm$9.04}  & 35.28\scriptsize{$\pm$20.12} & 18.18\scriptsize{$\pm$23.82} & 74.53\scriptsize{$\pm$1.87}  & 7.96\scriptsize{$\pm$14.23}  & 86.07\scriptsize{$\pm$0.36} & 78.91\scriptsize{$\pm$1.75} \\
QQP      & 86.67\scriptsize{$\pm$1.05} & 0.53\scriptsize{$\pm$1.66}  & 24.74\scriptsize{$\pm$10.03} & 0.00\scriptsize{$\pm$0.00}   & 23.23\scriptsize{$\pm$6.97}  & 0.00\scriptsize{$\pm$0.02}   & 88.25\scriptsize{$\pm$0.54} & 81.49\scriptsize{$\pm$1.47} \\
MRPC     & 80.75\scriptsize{$\pm$1.73} & 0.17\scriptsize{$\pm$0.30}  & 50.15\scriptsize{$\pm$19.65} & 0.01\scriptsize{$\pm$0.05}   & 49.15\scriptsize{$\pm$4.56}  & 0.01\scriptsize{$\pm$0.05}   & 86.03\scriptsize{$\pm$1.41} & 72.71\scriptsize{$\pm$2.82} \\
MNLI     & 81.39\scriptsize{$\pm$4.7}  & 0.78\scriptsize{$\pm$0.88}  & 12.90\scriptsize{$\pm$8.21}   & 0.87\scriptsize{$\pm$1.16}   & 57.30\scriptsize{$\pm$1.53}  & 0.00\scriptsize{$\pm$0.00}   & 83.50\scriptsize{$\pm$4.79}  & 76.71\scriptsize{$\pm$2.44} \\
QNLI     & 85.12\scriptsize{$\pm$5.57} & 0.05\scriptsize{$\pm$0.07}  & 27.76\scriptsize{$\pm$10.04} & 0.00\scriptsize{$\pm$0.00}   & 14.90\scriptsize{$\pm$8.48}  & 4.22\scriptsize{$\pm$5.46}   & 93.68\scriptsize{$\pm$0.41} & 77.53\scriptsize{$\pm$7.48} \\
RTE      & 84.24\scriptsize{$\pm$1.16} & 0.19\scriptsize{$\pm$0.77}  & 29.51\scriptsize{$\pm$15.12} & 0.00\scriptsize{$\pm$0.00}   & 47.67\scriptsize{$\pm$1.92}  & 3.16\scriptsize{$\pm$4.40}   & 93.26\scriptsize{$\pm$0.51} & 80.73\scriptsize{$\pm$3.24} \\
WNLI     & 62.34\scriptsize{$\pm$3.31} & 0.00\scriptsize{$\pm$0.00}  & 22.57\scriptsize{$\pm$15.96} & 0.00\scriptsize{$\pm$0.00}   & 41.08\scriptsize{$\pm$1.71}  & 3.62\scriptsize{$\pm$5.10}   & 77.53\scriptsize{$\pm$1.4}  & 61.07\scriptsize{$\pm$6.22} \\
MMLU     & 45.25\scriptsize{$\pm$0.83} & -                           & 15.31\scriptsize{$\pm$7.41}  & -                            & 36.05\scriptsize{$\pm$7.76}  & -                            & 53.04\scriptsize{$\pm$0.67} & 63.33\scriptsize{$\pm$2.56} \\
SQuAD V2 & 87.32\scriptsize{$\pm$0.43} & -                           & -                            & -                            & -                            & -                            & 89.78\scriptsize{$\pm$0.71} & 68.35\scriptsize{$\pm$4.36} \\
IWSLT    & 0.18\scriptsize{$\pm$0.04}  & -                           & -                            & -                            & -                            & -                            & 0.21\scriptsize{$\pm$0.04}  & 0.23\scriptsize{$\pm$0.01}  \\
UN Multi & 0.29\scriptsize{$\pm$0.02}  & -                           & -                            & -                            & -                            & -                            & 0.33\scriptsize{$\pm$0.02}  & 0.34\scriptsize{$\pm$0.01}  \\
Math     & 14.22\scriptsize{$\pm$3.25} & -                           & -                            & -                            & -                            & -                            & 14.81\scriptsize{$\pm$1.35} & 13.14\scriptsize{$\pm$8.48} \\

\bottomrule

\end{tabular}
}
% \vspace{-.2in}
\end{table}

\subsection{Analysis on model size and fine-tuning}
\label{sec-append-size-finetune}

\begin{table*}[t!]
\centering
\caption{The APDR and standard deviations of different attacks on different models.}
\label{tb-model-attack}
\resizebox{.8\textwidth}{!}{
\begin{tabular}{lccccccc}
\toprule

\multirow{2}{*}{Model} & \multicolumn{2}{c}{Character-level} & \multicolumn{2}{c}{Word-level} & \multicolumn{2}{c}{Sentence-level} & \multicolumn{1}{c}{Semantic-level} \\
\cmidrule(lr){2-3} \cmidrule(lr){4-5} \cmidrule(lr){6-7} \cmidrule(lr){8-8}
& TextBugger & DeepWordBug & TextFooler & BertAttack & CheckList & StressTest & Semantic \\
\midrule

T5-large        & 0.09\scriptsize{$\pm$0.10} & 0.13\scriptsize{$\pm$0.18} & 0.20\scriptsize{$\pm$0.24} & 0.21\scriptsize{$\pm$0.24} & 0.04\scriptsize{$\pm$0.08} & 0.18\scriptsize{$\pm$0.24} & 0.10\scriptsize{$\pm$0.09} \\
Vicuna          & 0.81\scriptsize{$\pm$0.25} & 0.69\scriptsize{$\pm$0.30} & 0.80\scriptsize{$\pm$0.26} & 0.84\scriptsize{$\pm$0.23} & 0.64\scriptsize{$\pm$0.27} & 0.29\scriptsize{$\pm$0.40} & 0.74\scriptsize{$\pm$0.25} \\
Llama2          & 0.67\scriptsize{$\pm$0.36} & 0.41\scriptsize{$\pm$0.34} & 0.68\scriptsize{$\pm$0.36} & 0.74\scriptsize{$\pm$0.33} & 0.34\scriptsize{$\pm$0.33} & 0.20\scriptsize{$\pm$0.30} & 0.66\scriptsize{$\pm$0.35} \\
UL2             & 0.04\scriptsize{$\pm$0.06} & 0.03\scriptsize{$\pm$0.04} & 0.14\scriptsize{$\pm$0.20} & 0.16\scriptsize{$\pm$0.22} & 0.04\scriptsize{$\pm$0.07} & 0.06\scriptsize{$\pm$0.09} & 0.06\scriptsize{$\pm$0.08} \\
ChatGPT         & 0.14\scriptsize{$\pm$0.20} & 0.08\scriptsize{$\pm$0.13} & 0.32\scriptsize{$\pm$0.35} & 0.34\scriptsize{$\pm$0.34} & 0.07\scriptsize{$\pm$0.13} & 0.06\scriptsize{$\pm$0.12} & 0.26\scriptsize{$\pm$0.22} \\
GPT-4            & 0.03\scriptsize{$\pm$0.10} & 0.02\scriptsize{$\pm$0.08} & 0.18\scriptsize{$\pm$0.19} & 0.27\scriptsize{$\pm$0.40} & -0.02\scriptsize{$\pm$0.09} & 0.03\scriptsize{$\pm$0.15} & 0.03\scriptsize{$\pm$0.16} \\
\midrule
Avg      & 0.21\scriptsize{$\pm$0.30} & 0.17\scriptsize{$\pm$0.26} & 0.31\scriptsize{$\pm$0.33} & 0.33\scriptsize{$\pm$0.34} & 0.12\scriptsize{$\pm$0.23} & 0.11\scriptsize{$\pm$0.23}  & 0.22\scriptsize{$\pm$0.26}
\\
\bottomrule

\end{tabular}
}
\end{table*}

As shown in \cref{tb-model-attack} and \ref{tb-dataset-model}, there seems to be no clear correlation between model robustness and size, for example, despite being the smallest, T5-large demonstrates robustness on par with larger models such as \chat on our evaluated datasets. 

The observed differences in model robustness might stem from two aspects: 1) the specific fine-tuning techniques employed. For example, both UL2 and T5-large, fine-tuned on large datasets, and \chat, fine-tuned via RLHF~\citep{christiano2017deep}, exhibit better robustness than Vicuna. These findings encourage further investigation of fine-tuning strategies to enhance robustness. 2) the memorization of training data. Recent studies suggest that the remarkable performance of some \llms might be rooted in their ability to memorize training data \citep{stochasticparrot, magar2022data, carlini2023quantifying, biderman2023emergent}, rather than in generalization. Hence, even when confronted with adversarial prompts, models might leverage this memorization to produce accurate responses.

In this section, we conduct experiments to analyze the effects of different model sizes and fine-tuning on adversarial prompts.
Particularly, we leverage the open-source Llama2~\citep{touvron2023llama2} series due to their support on different sizes and their corresponding fine-tuned versions.
The chat versions of Llama2 (Llama2-chat) are fine-tuned on human instructions datasets to better follow the instructions and support multi-turn conversations, while the original version can only be used for inference.

\begin{figure}[htbp]
\centering

\subfigure[Analysis of model size.]{
    \includegraphics[width=0.6\textwidth]{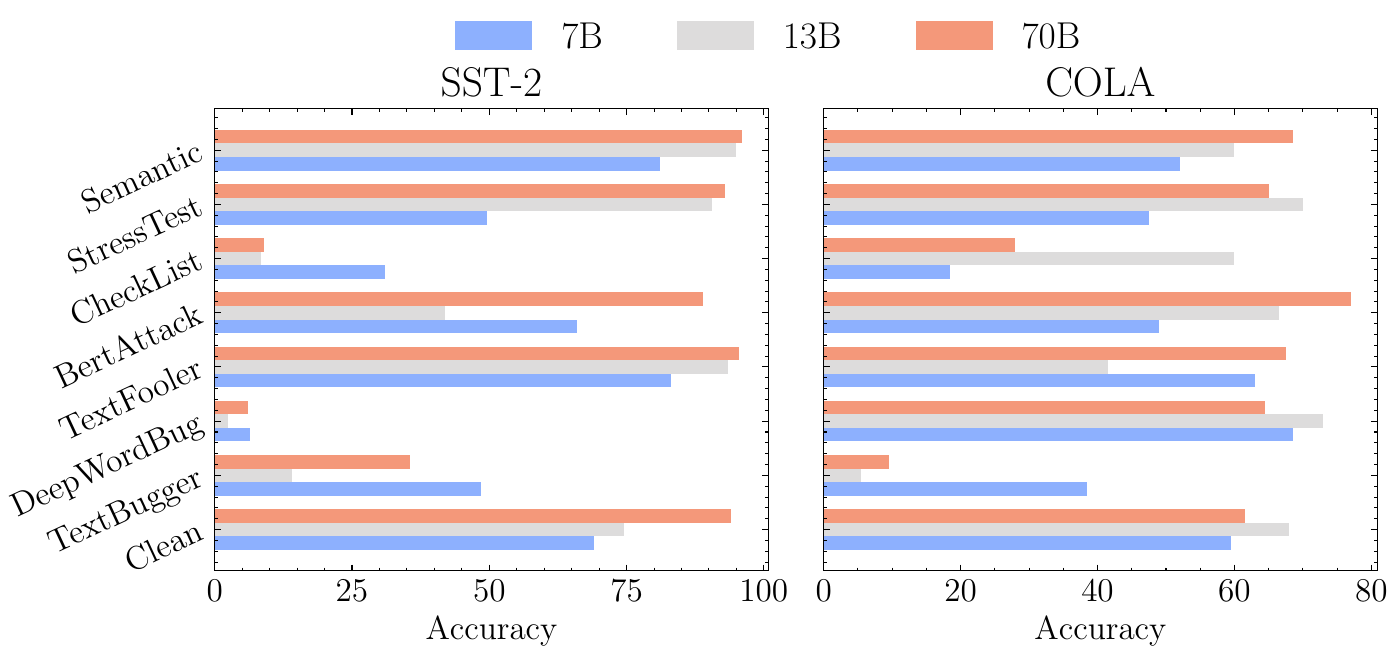}
    \label{fig-model-size}
}
\subfigure[Analysis of fine-tuning.]{
    \includegraphics[width=0.6\textwidth]{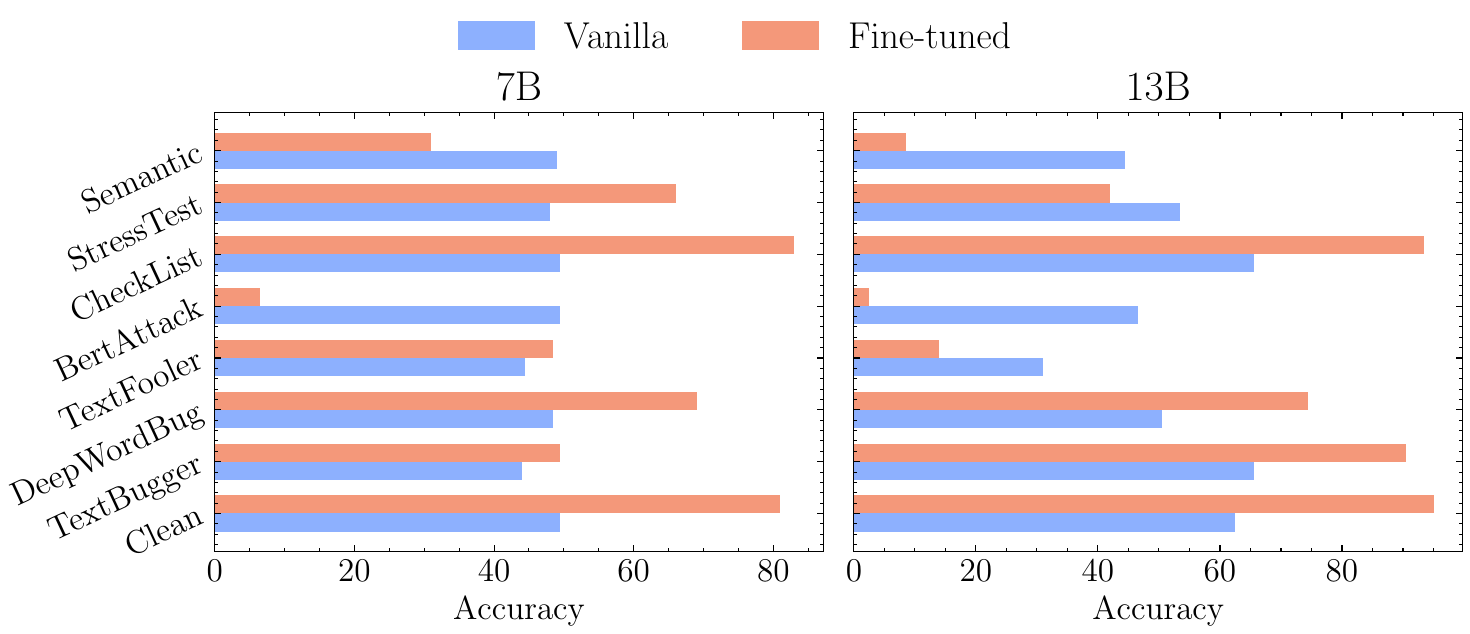}
    \label{fig-finetune}
}

\caption{(a) Accuracy of Llama2 models (7B-chat, 13B-chat, 70B-chat) on SST2 and CoLA datasets. (b) Accuracy of Llama2 models with fine-tuning and w/o fine-tuning (vanilla) on SST-2 dataset.}
\label{fig-size-finetune}
\end{figure}

\textbf{Robustness of different model sizes}
We analyzed three models from the open-source Llama2 series~\citep{touvron2023llama2}: Llama2-7B-chat, Llama2-13B-chat, and Llama2-70B-chat. These were chosen due to their distinct sizes, further, their fine-tuning datasets and methods are the same. Our results, depicted in \figurename~\ref{fig-model-size}, reveal that, in a non-adversarial setting, larger models like the 70B model typically surpass the performance of their smaller counterparts. Yet, when subjected to adversarial attacks, the performance dynamics change: at times, smaller models outshine the larger ones. 
The reasons for these abnormal behaviors could trigger interests for future research. 

% This anomaly might stem from the amplified effects of gradient accumulation in the larger models, a phenomenon further discussed in Sec. \ref{sec-exp-visualzation}. \zkj{Do not understand the meaning of 'gradient accumulation'}

\textbf{Robustness of fine-tuning}
To delve into the intricacies of fine-tuning, we compared the performances of Llama2-7B and Llama2-7B-chat on the SST2 and COLA tasks. Our analysis, as visualized in \figurename~\ref{fig-finetune}, underscores a consistent trend: models fine-tuned using human-instruction datasets fare better against adversarial onslaughts than models that are not fine-tuned.
This observation implies that fine-tuning could be further utilized as the countermeasures for adversarial inputs.
\begin{table}[htbp]
\centering
\caption{Accuracy (\%) of GPT-4 on clean and adversarial prompts and samples on the AdvGLUE \cite{wang2021adversarial} dataset, i.e., attacking both prompts and samples.}
\label{tb-advglue-all}
\resizebox{.6\textwidth}{!}{
\begin{tabular}{l|ccccc|c}
\toprule
Attack & SST-2 & QQP & MNLI & QNLI & RTE & AVG \\ \midrule
\makecell{Clean prompts \\\& clean samples} & 96.10 & 78.23 & 81.05 & 64.50 & 87.54 & 81.48 \\ \midrule 
\makecell{Clean prompts \\\& AdvGLUE} & 63.51 & 70.51 & 63.64 & 62.84 & 74.07 & 66.91 \\ \midrule
TextBugger & 58.78 & 44.87 & 47.93 & 60.81 & 76.54 & 57.79 \\ 
DeepWordBug & 66.22 & 61.54 & 59.50 & 61.49 & 72.84 & 64.32 \\ 
TextFooler & 2.03 & 1.28 & 46.28 & 4.05 & 0.00 & 10.73 \\ 
BertAttack & 0.00 & 0.00 & 27.27 & 24.32 & 71.60 & 24.64 \\ 
Checklist & 69.59 & 66.67 & 57.85 & 56.08 & 72.84 & 64.61 \\ 
StressTest & 50.68 & 56.41 & 59.50 & 59.46 & 76.54 & 60.52 \\ 
Semantic & 0.00 & 38.46 & 48.76 & 65.54 & 41.98 & 38.95 \\ \bottomrule
\end{tabular}
}
\end{table}

\subsection{Results excluding non-semantic preserving adversarial prompts}
\label{appendix-excluded-results}

\tablename~\ref{tb-append-dataset-attack} presents the attack results after excluding adversarial prompts that do not preserve semantics. It can be observed that the APDR is still considerably high for each dataset.

\begin{table*}[t!]
\centering
\caption{The APDR and standard deviations of different attacks on different datasets by \emph{excluding} the ones human annotators do not think acceptable.}
% \vspace{-.1in}
\label{tb-append-dataset-attack}
\resizebox{.9\textwidth}{!}{
\begin{tabular}{lccccccc}
\toprule

\multirow{2}{*}{Dataset} & \multicolumn{2}{c}{Character-level} & \multicolumn{2}{c}{Word-level} & \multicolumn{2}{c}{Sentence-level} & \multicolumn{1}{c}{Semantic-level} \\
\cmidrule(lr){2-3} \cmidrule(lr){4-5} \cmidrule(lr){6-7} \cmidrule(lr){8-8}
& TextBugger & DeepWordBug & TextFooler & BertAttack & CheckList & StressTest & Semantic \\
\midrule
SST-2    & 0.26\scriptsize{$\pm$0.39} & 0.21\scriptsize{$\pm$0.36} & {\color{black}0.33\scriptsize{$\pm$0.35}} & {\color{black}0.30\scriptsize{$\pm$0.39}} & 0.27\scriptsize{$\pm$0.39} & 0.17\scriptsize{$\pm$0.34}  & 0.28\scriptsize{$\pm$0.36} \\
CoLA     & 0.37\scriptsize{$\pm$0.39} & 0.29\scriptsize{$\pm$0.36} & {\color{black}0.40\scriptsize{$\pm$0.33}} & {\color{black}0.42\scriptsize{$\pm$0.31}} & 0.25\scriptsize{$\pm$0.32} & 0.21\scriptsize{$\pm$0.28}  & 0.27\scriptsize{$\pm$0.35} \\
QQP      & 0.20\scriptsize{$\pm$0.32} & 0.18\scriptsize{$\pm$0.27} & {\color{black}0.26\scriptsize{$\pm$0.31}} & {\color{black}0.29\scriptsize{$\pm$0.33}} & 0.13\scriptsize{$\pm$0.25} & -0.00\scriptsize{$\pm$0.21} & 0.30\scriptsize{$\pm$0.36} \\
MRPC     & 0.24\scriptsize{$\pm$0.33} & 0.21\scriptsize{$\pm$0.30} & {\color{black}0.27\scriptsize{$\pm$0.30}} & {\color{black}0.31\scriptsize{$\pm$0.29}} & 0.13\scriptsize{$\pm$0.27} & 0.20\scriptsize{$\pm$0.30}  & 0.28\scriptsize{$\pm$0.36} \\
MNLI     & 0.26\scriptsize{$\pm$0.37} & 0.18\scriptsize{$\pm$0.31} & {\color{black}0.27\scriptsize{$\pm$0.36}} & {\color{black}0.34\scriptsize{$\pm$0.32}} & 0.16\scriptsize{$\pm$0.26} & 0.11\scriptsize{$\pm$0.27}  & 0.11\scriptsize{$\pm$0.04} \\
QNLI     & 0.36\scriptsize{$\pm$0.39} & 0.41\scriptsize{$\pm$0.36} & {\color{black}0.47\scriptsize{$\pm$0.33}} & {\color{black}0.45\scriptsize{$\pm$0.30}} & 0.22\scriptsize{$\pm$0.37} & 0.18\scriptsize{$\pm$0.26}  & 0.35\scriptsize{$\pm$0.33} \\
RTE      & 0.24\scriptsize{$\pm$0.37} & 0.22\scriptsize{$\pm$0.36} & {\color{black}0.26\scriptsize{$\pm$0.34}} & {\color{black}0.28\scriptsize{$\pm$0.35}} & 0.19\scriptsize{$\pm$0.32} & 0.18\scriptsize{$\pm$0.25}  & 0.28\scriptsize{$\pm$0.33} \\
WNLI     & 0.28\scriptsize{$\pm$0.36} & 0.26\scriptsize{$\pm$0.35} & {\color{black}0.27\scriptsize{$\pm$0.31}} & {\color{black}0.28\scriptsize{$\pm$0.29}} & 0.19\scriptsize{$\pm$0.30} & 0.19\scriptsize{$\pm$0.26}  & 0.36\scriptsize{$\pm$0.32} \\
MMLU     & 0.18\scriptsize{$\pm$0.22} & 0.11\scriptsize{$\pm$0.15} & {\color{black}0.19\scriptsize{$\pm$0.16}} & {\color{black}0.31\scriptsize{$\pm$0.20}} & 0.14\scriptsize{$\pm$0.20} & 0.03\scriptsize{$\pm$0.16}  & 0.17\scriptsize{$\pm$0.17} \\
SQuAD V2 & 0.09\scriptsize{$\pm$0.17} & 0.05\scriptsize{$\pm$0.08} & {\color{black}0.23\scriptsize{$\pm$0.25}} & {\color{black}0.30\scriptsize{$\pm$0.29}} & 0.02\scriptsize{$\pm$0.03} & 0.02\scriptsize{$\pm$0.04}  & 0.07\scriptsize{$\pm$0.09} \\
IWSLT    & 0.09\scriptsize{$\pm$0.14} & 0.11\scriptsize{$\pm$0.12} & {\color{black}0.26\scriptsize{$\pm$0.25}} & {\color{black}0.12\scriptsize{$\pm$0.16}} & 0.10\scriptsize{$\pm$0.10} & 0.17\scriptsize{$\pm$0.19}  & 0.18\scriptsize{$\pm$0.14} \\
UN Multi & 0.06\scriptsize{$\pm$0.08} & 0.08\scriptsize{$\pm$0.12} & {\color{black}0.17\scriptsize{$\pm$0.19}} & {\color{black}0.10\scriptsize{$\pm$0.13}} & 0.06\scriptsize{$\pm$0.07} & 0.09\scriptsize{$\pm$0.11}  & 0.15\scriptsize{$\pm$0.18} \\
Math     & 0.19\scriptsize{$\pm$0.17} & 0.15\scriptsize{$\pm$0.13} & {\color{black}0.45\scriptsize{$\pm$0.32}} & {\color{black}0.39\scriptsize{$\pm$0.27}} & 0.16\scriptsize{$\pm$0.11} & 0.13\scriptsize{$\pm$0.08}  & 0.23\scriptsize{$\pm$0.13} \\
\midrule
Avg      & 0.23\scriptsize{$\pm$0.33} & 0.20\scriptsize{$\pm$0.30} & {\color{black}0.29\scriptsize{$\pm$0.31}} & {\color{black}0.31\scriptsize{$\pm$0.30}} & 0.16\scriptsize{$\pm$0.27} & 0.13\scriptsize{$\pm$0.25}  & 0.24\scriptsize{$\pm$0.29}
\\
\bottomrule

\end{tabular}
}
% \vspace{-.2in}
\end{table*}

\subsection{Attacking both prompts and samples}
\label{sec-exp-advglue}

The primary focus of this work is to evaluate the robustness of prompts rather than input samples since the samples can be omitted in certain situations, as discussed in Sec.~\ref{sec-attack}.
In this section, we explore the possibility of attacking both prompts and samples, i.e., evaluating the performance of \llms in adversarial prompts and samples.
Note that since the generation of adversarial examples is expensive and time-consuming, we leverage an existing adversarial dataset called AdvGLUE \citep{wang2021adversarial}, which contains adversarial examples from GLUE \citep{wang2019glue} and it consists of five same tasks as GLUE: SST-2, QQP, MNLI, QNLI, and RTE.
Then, we leverage the adversarial prompts and the AdvGLUE dataset \citep{wang2021adversarial} to evaluate the performance when attacking both prompts and samples.

\tablename~\ref{tb-advglue} shows the accuracy results using both clean and adversarial prompts on AdvGLUE and clean dataset, respectively.
The results demonstrate that on average, all attacking approaches are effective since the accuracy is dramatically declined in face of adversarial prompts.
Similar to \tablename~\ref{tb-dataset-attack}, word-level attacks (TextFooler and BertAttack) are the most effective with more than 49\% of accuracy drop.
Moreover, surprising results emerge for Checklist attack since the performance can sometimes be improved (e.g., 69.59\% on SST-2 vs. the original 63.51\%).
This is also consistent with our previous observation in Sec.~\ref{stresstest analysis}.
The results in this section show that attacking both the prompts and samples can further reduce the performance of \llms.
However, certain attacks can even enhance the performance, which is left for future research.

\section{Effectiveness Analysis of LLM Response and Attention Visualization}
\label{sec-append-analysis}

\subsection{Erroneous response analysis}
\label{sec-append-error}

We first analyze the erroneous response analysis produced by adversarial prompts. The results suggest that adversarial prompts impact \llms' performance by inducing misclassification errors and hindering their ability to generate meaningful responses. 

\begin{itemize}

\item \textbf{Induced Misclassification:} As exemplified by BertAttack, CheckList, and Translation attacks, adversarial prompts can lead the model to erroneous classifications. For instance, the sentiment prediction may shift from positive to negative due to the influence of the adversarial prompt. This instance validates the efficacy of adversarial attacks in manipulating the model's decision-making processes.

\item \textbf{Generation of Incoherent Responses:} In the case of the DeepWordBug attack, the adversarial prompt results in the model generating incoherent or nonsensical sentences. For example, the response ``\prompt{None of the above choices}'' does not align with any positive or negative sentiment classification, thereby demonstrating that the model fails to comprehend the intended input. This observation emphasizes the susceptibility of LLMs to adversarial perturbations that can potentially hamper their natural language understanding capabilities.

\end{itemize}

\subsection{Attention visualization techniques}
\label{sec-append-attention}

\begin{table*}[t!]
\caption{Attention visualization of samples that are \textit{correctly classified by adv. prompt but misclassified by clean prompt.} Notations and colors follow \tablename~\ref{table: attention misclassified adv prompts}.
}
\label{table: attention correctly classified adv prompts}
\centering
\resizebox{\textwidth}{!}{
\begin{tabular}{l c m{18.5cm}}
\toprule
\multicolumn{1}{c}{\textbf{Attack}} & \multicolumn{1}{c}{\textbf{Pred.}} & \multicolumn{1}{c}{\textbf{[Prompt, sample]}} \\
\midrule

\multirow{4}{*}{\footnotesize CheckList} & {\color{red} E} & \scriptsize \colorbox[RGB]{254,226,213}{Determine\vphantom{fg}}\hspace*{0pt}\colorbox[RGB]{254,227,215}{if\vphantom{fg}}\hspace*{0pt}\colorbox[RGB]{255,245,240}{the\vphantom{fg}}\hspace*{0pt}\colorbox[RGB]{254,241,234}{given\vphantom{fg}}\hspace*{0pt}\colorbox[RGB]{254,227,214}{pair\vphantom{fg}}\hspace*{0pt}\colorbox[RGB]{254,242,236}{of\vphantom{fg}}\hspace*{0pt}\colorbox[RGB]{252,168,139}{statements\vphantom{fg}}\hspace*{0pt}\colorbox[RGB]{254,239,231}{can\vphantom{fg}}\hspace*{0pt}\colorbox[RGB]{255,245,240}{be\vphantom{fg}}\hspace*{0pt}\colorbox[RGB]{254,236,227}{considered\vphantom{fg}}\hspace*{0pt}\colorbox[RGB]{254,240,233}{the\vphantom{fg}}\hspace*{0pt}\colorbox[RGB]{252,195,172}{same\vphantom{fg}}\hspace*{0pt}\colorbox[RGB]{254,228,216}{by\vphantom{fg}}\hspace*{0pt}\colorbox[RGB]{252,191,166}{responding\vphantom{fg}}\hspace*{0pt}\colorbox[RGB]{254,237,229}{with\vphantom{fg}}\hspace*{0pt}\colorbox[RGB]{110,1,14}{'equivalent'\vphantom{fg}}\hspace*{0pt}\colorbox[RGB]{254,234,224}{or\vphantom{fg}}\hspace*{0pt}\colorbox[RGB]{244,80,57}{'not\_equivalent'.\vphantom{fg}}\hspace*{0pt}\colorbox[RGB]{253,222,208}{Question\vphantom{fg}}\hspace*{0pt}\colorbox[RGB]{252,157,126}{1:\vphantom{fg}}\hspace*{0pt}\colorbox[RGB]{252,149,117}{What\vphantom{fg}}\hspace*{0pt}\colorbox[RGB]{251,110,78}{language\vphantom{fg}}\hspace*{0pt}\colorbox[RGB]{251,131,99}{is\vphantom{fg}}

% \hspace*{0pt}

\colorbox[RGB]{103,0,12}{this?\vphantom{fg}}\hspace*{0pt}\colorbox[RGB]{254,228,216}{Question\vphantom{fg}}\hspace*{0pt}\colorbox[RGB]{252,149,117}{2:\vphantom{fg}}\hspace*{0pt}\colorbox[RGB]{252,155,125}{What\vphantom{fg}}\hspace*{0pt}\colorbox[RGB]{252,171,142}{language\vphantom{fg}}\hspace*{0pt}\colorbox[RGB]{252,171,142}{is\vphantom{fg}}\hspace*{0pt}\colorbox[RGB]{223,44,37}{this\vphantom{fg}}\hspace*{0pt}\colorbox[RGB]{246,87,62}{in?\vphantom{fg}}\hspace*{0pt}\colorbox[RGB]{254,234,224}{Answer:\vphantom{fg}}\hspace*{0pt}
\\

\cmidrule{2-3}

& {\footnotesize \color{green}{NE}} & \scriptsize \colorbox[RGB]{254,231,221}{Determine\vphantom{fg}}\hspace*{0pt}\colorbox[RGB]{254,229,217}{if\vphantom{fg}}\hspace*{0pt}\colorbox[RGB]{254,237,229}{the\vphantom{fg}}\hspace*{0pt}\colorbox[RGB]{254,235,225}{given\vphantom{fg}}\hspace*{0pt}\colorbox[RGB]{253,219,203}{pair\vphantom{fg}}\hspace*{0pt}\colorbox[RGB]{254,241,235}{of\vphantom{fg}}\hspace*{0pt}\colorbox[RGB]{252,175,147}{statements\vphantom{fg}}\hspace*{0pt}\colorbox[RGB]{254,240,233}{can\vphantom{fg}}\hspace*{0pt}\colorbox[RGB]{255,245,240}{be\vphantom{fg}}\hspace*{0pt}\colorbox[RGB]{254,234,224}{considered\vphantom{fg}}\hspace*{0pt}\colorbox[RGB]{254,237,228}{the\vphantom{fg}}\hspace*{0pt}\colorbox[RGB]{252,186,160}{same\vphantom{fg}}\hspace*{0pt}\colorbox[RGB]{254,233,223}{by\vphantom{fg}}\hspace*{0pt}\colorbox[RGB]{253,213,195}{responding\vphantom{fg}}\hspace*{0pt}\colorbox[RGB]{254,229,217}{with\vphantom{fg}}\hspace*{0pt}\colorbox[RGB]{178,18,23}{'equivalent'\vphantom{fg}}\hspace*{0pt}\colorbox[RGB]{254,228,216}{or\vphantom{fg}}\hspace*{0pt}\colorbox[RGB]{188,20,26}{'not\_equivalent'\vphantom{fg}}\hspace*{0pt}\colorbox[RGB]{103,0,12}{EAB4KP2NVY.\vphantom{fg}}\hspace*{0pt}\colorbox[RGB]{254,231,220}{Question\vphantom{fg}}\hspace*{0pt}\colorbox[RGB]{252,160,131}{1:\vphantom{fg}}\hspace*{0pt}\colorbox[RGB]{252,192,168}{What\vphantom{fg}}

% \hspace*{0pt}

\colorbox[RGB]{252,184,157}{language\vphantom{fg}}\hspace*{0pt}\colorbox[RGB]{252,205,185}{is\vphantom{fg}}\hspace*{0pt}\colorbox[RGB]{243,77,55}{this?\vphantom{fg}}\hspace*{0pt}\colorbox[RGB]{254,230,219}{Question\vphantom{fg}}\hspace*{0pt}\colorbox[RGB]{252,163,134}{2:\vphantom{fg}}\hspace*{0pt}\colorbox[RGB]{253,208,189}{What\vphantom{fg}}\hspace*{0pt}\colorbox[RGB]{252,205,185}{language\vphantom{fg}}\hspace*{0pt}\colorbox[RGB]{254,225,211}{is\vphantom{fg}}\hspace*{0pt}\colorbox[RGB]{252,153,122}{this\vphantom{fg}}\hspace*{0pt}\colorbox[RGB]{252,168,139}{in?\vphantom{fg}}\hspace*{0pt}\colorbox[RGB]{254,235,225}{Answer:\vphantom{fg}}\hspace*{0pt}

\\
\midrule

\multirow{6}{*}{\footnotesize StressTest}  & {\footnotesize {\color{red}E}} & \scriptsize  \colorbox[RGB]{254,244,239}{As\vphantom{fg}}\hspace*{0pt}\colorbox[RGB]{255,245,240}{an\vphantom{fg}}\hspace*{0pt}\colorbox[RGB]{254,243,238}{instrument\vphantom{fg}}\hspace*{0pt}\colorbox[RGB]{254,244,239}{for\vphantom{fg}}\hspace*{0pt}\colorbox[RGB]{253,211,192}{entailment\vphantom{fg}}\hspace*{0pt}\colorbox[RGB]{254,237,228}{evaluation,\vphantom{fg}}\hspace*{0pt}\colorbox[RGB]{255,245,240}{consider\vphantom{fg}}\hspace*{0pt}\colorbox[RGB]{254,240,233}{the\vphantom{fg}}\hspace*{0pt}\colorbox[RGB]{254,243,237}{two\vphantom{fg}}\hspace*{0pt}\colorbox[RGB]{254,239,232}{sentences\vphantom{fg}}\hspace*{0pt}\colorbox[RGB]{254,242,236}{and\vphantom{fg}}\hspace*{0pt}\colorbox[RGB]{255,245,240}{determine\vphantom{fg}}\hspace*{0pt}\colorbox[RGB]{254,241,235}{if\vphantom{fg}}\hspace*{0pt}\colorbox[RGB]{254,243,238}{their\vphantom{fg}}\hspace*{0pt}\colorbox[RGB]{254,240,233}{relationship\vphantom{fg}}\hspace*{0pt}\colorbox[RGB]{254,241,234}{is\vphantom{fg}}\hspace*{0pt}\colorbox[RGB]{252,205,185}{'entailment'\vphantom{fg}}\hspace*{0pt}\colorbox[RGB]{254,241,234}{or\vphantom{fg}}\hspace*{0pt}\colorbox[RGB]{252,197,174}{'not\_entailment'.\vphantom{fg}}\hspace*{0pt}\colorbox[RGB]{254,237,229}{Respond\vphantom{fg}}\hspace*{0pt}\colorbox[RGB]{254,239,232}{with\vphantom{fg}}

% \hspace*{0pt}

\colorbox[RGB]{252,191,166}{'entailment'\vphantom{fg}}\hspace*{0pt}\colorbox[RGB]{254,239,232}{or\vphantom{fg}}\hspace*{0pt}\colorbox[RGB]{252,198,175}{'not\_entailment'\vphantom{fg}}\hspace*{0pt}\colorbox[RGB]{254,225,212}{:\vphantom{fg}}\hspace*{0pt}\colorbox[RGB]{254,231,221}{Sentence\vphantom{fg}}\hspace*{0pt}\colorbox[RGB]{254,229,217}{1:\vphantom{fg}}\hspace*{0pt}\colorbox[RGB]{253,207,188}{Look!\vphantom{fg}}\hspace*{0pt}\colorbox[RGB]{253,220,205}{There\vphantom{fg}}\hspace*{0pt}\colorbox[RGB]{252,199,177}{is\vphantom{fg}}\hspace*{0pt}\colorbox[RGB]{252,175,147}{a\vphantom{fg}}\hspace*{0pt}\colorbox[RGB]{103,0,12}{minnow\vphantom{fg}}\hspace*{0pt}\colorbox[RGB]{251,139,107}{swimming\vphantom{fg}}\hspace*{0pt}\colorbox[RGB]{252,168,139}{right\vphantom{fg}}\hspace*{0pt}\colorbox[RGB]{243,77,55}{below\vphantom{fg}}\hspace*{0pt}\colorbox[RGB]{251,124,92}{that\vphantom{fg}}\hspace*{0pt}\colorbox[RGB]{245,86,61}{duck!\vphantom{fg}}\hspace*{0pt}\colorbox[RGB]{251,127,95}{It\vphantom{fg}}\hspace*{0pt}\colorbox[RGB]{252,180,153}{had\vphantom{fg}}\hspace*{0pt}\colorbox[RGB]{252,171,142}{better\vphantom{fg}}\hspace*{0pt}\colorbox[RGB]{252,202,182}{get\vphantom{fg}}\hspace*{0pt}\colorbox[RGB]{252,198,175}{away\vphantom{fg}}\hspace*{0pt}\colorbox[RGB]{254,225,212}{to\vphantom{fg}}\hspace*{0pt}\colorbox[RGB]{253,214,197}{safety\vphantom{fg}}\hspace*{0pt}\colorbox[RGB]{252,201,180}{fast!\vphantom{fg}}

% \hspace*{0pt}

\colorbox[RGB]{253,221,206}{Sentence\vphantom{fg}}\hspace*{0pt}\colorbox[RGB]{253,211,192}{2:\vphantom{fg}}\hspace*{0pt}\colorbox[RGB]{254,227,214}{The\vphantom{fg}}\hspace*{0pt}\colorbox[RGB]{252,168,139}{duck\vphantom{fg}}\hspace*{0pt}\colorbox[RGB]{252,202,182}{had\vphantom{fg}}\hspace*{0pt}\colorbox[RGB]{252,199,177}{better\vphantom{fg}}\hspace*{0pt}\colorbox[RGB]{254,224,210}{get\vphantom{fg}}\hspace*{0pt}\colorbox[RGB]{253,223,209}{away\vphantom{fg}}\hspace*{0pt}\colorbox[RGB]{254,233,223}{to\vphantom{fg}}\hspace*{0pt}\colorbox[RGB]{254,231,221}{safety\vphantom{fg}}\hspace*{0pt}\colorbox[RGB]{253,218,202}{fast!\vphantom{fg}}\hspace*{0pt}\colorbox[RGB]{254,234,224}{Answer:\vphantom{fg}}\hspace*{0pt}

\\
\cmidrule{2-3}
& {\footnotesize {\color{green} NE}} & \scriptsize \colorbox[RGB]{255,245,240}{As\vphantom{fg}}\hspace*{0pt}\colorbox[RGB]{255,245,240}{an\vphantom{fg}}\hspace*{0pt}\colorbox[RGB]{254,244,239}{instrument\vphantom{fg}}\hspace*{0pt}\colorbox[RGB]{255,245,240}{for\vphantom{fg}}\hspace*{0pt}\colorbox[RGB]{253,211,192}{entailment\vphantom{fg}}\hspace*{0pt}\colorbox[RGB]{254,237,229}{evaluation,\vphantom{fg}}\hspace*{0pt}\colorbox[RGB]{255,245,240}{consider\vphantom{fg}}\hspace*{0pt}\colorbox[RGB]{254,241,235}{the\vphantom{fg}}\hspace*{0pt}\colorbox[RGB]{254,243,237}{two\vphantom{fg}}\hspace*{0pt}\colorbox[RGB]{254,240,233}{sentences\vphantom{fg}}\hspace*{0pt}\colorbox[RGB]{254,243,238}{and\vphantom{fg}}\hspace*{0pt}\colorbox[RGB]{254,244,239}{determine\vphantom{fg}}\hspace*{0pt}\colorbox[RGB]{254,242,236}{if\vphantom{fg}}\hspace*{0pt}\colorbox[RGB]{254,244,239}{their\vphantom{fg}}\hspace*{0pt}\colorbox[RGB]{254,241,234}{relationship\vphantom{fg}}\hspace*{0pt}\colorbox[RGB]{254,241,234}{is\vphantom{fg}}\hspace*{0pt}\colorbox[RGB]{252,202,182}{'entailment'\vphantom{fg}}\hspace*{0pt}\colorbox[RGB]{254,241,234}{or\vphantom{fg}}\hspace*{0pt}\colorbox[RGB]{252,193,169}{'not\_entailment'.\vphantom{fg}}\hspace*{0pt}\colorbox[RGB]{254,239,231}{Respond\vphantom{fg}}\hspace*{0pt}\colorbox[RGB]{254,240,233}{with\vphantom{fg}}

% \hspace*{0pt}

\colorbox[RGB]{252,180,153}{'entailment'\vphantom{fg}}\hspace*{0pt}\colorbox[RGB]{254,239,231}{or\vphantom{fg}}\hspace*{0pt}\colorbox[RGB]{252,150,119}{'not\_entailment'\vphantom{fg}}\hspace*{0pt}\colorbox[RGB]{254,231,221}{and\vphantom{fg}}\hspace*{0pt}\colorbox[RGB]{254,235,226}{true\vphantom{fg}}\hspace*{0pt}\colorbox[RGB]{254,235,226}{is\vphantom{fg}}\hspace*{0pt}\colorbox[RGB]{254,238,230}{true\vphantom{fg}}\hspace*{0pt}\colorbox[RGB]{254,227,214}{:Sentence\vphantom{fg}}\hspace*{0pt}\colorbox[RGB]{254,230,219}{1:\vphantom{fg}}\hspace*{0pt}\colorbox[RGB]{253,207,188}{Look!\vphantom{fg}}\hspace*{0pt}\colorbox[RGB]{253,220,205}{There\vphantom{fg}}\hspace*{0pt}\colorbox[RGB]{252,198,175}{is\vphantom{fg}}\hspace*{0pt}\colorbox[RGB]{252,175,147}{a\vphantom{fg}}\hspace*{0pt}\colorbox[RGB]{103,0,12}{minnow\vphantom{fg}}\hspace*{0pt}\colorbox[RGB]{251,141,109}{swimming\vphantom{fg}}\hspace*{0pt}\colorbox[RGB]{252,169,141}{right\vphantom{fg}}\hspace*{0pt}\colorbox[RGB]{244,81,58}{below\vphantom{fg}}\hspace*{0pt}\colorbox[RGB]{251,126,94}{that\vphantom{fg}}\hspace*{0pt}\colorbox[RGB]{247,90,64}{duck!\vphantom{fg}}\hspace*{0pt}\colorbox[RGB]{251,127,95}{It\vphantom{fg}}\hspace*{0pt}\colorbox[RGB]{252,180,153}{had\vphantom{fg}}\hspace*{0pt}\colorbox[RGB]{252,171,142}{better\vphantom{fg}}\hspace*{0pt}\colorbox[RGB]{252,205,185}{get\vphantom{fg}}\hspace*{0pt}\colorbox[RGB]{252,199,177}{away\vphantom{fg}}\hspace*{0pt}\colorbox[RGB]{254,225,212}{to\vphantom{fg}}\hspace*{0pt}\colorbox[RGB]{253,214,197}{safety\vphantom{fg}}

% \hspace*{0pt}

\colorbox[RGB]{252,202,182}{fast!\vphantom{fg}}\hspace*{0pt}\colorbox[RGB]{254,224,210}{Sentence\vphantom{fg}}\hspace*{0pt}\colorbox[RGB]{253,214,197}{2:\vphantom{fg}}\hspace*{0pt}\colorbox[RGB]{254,227,214}{The\vphantom{fg}}\hspace*{0pt}\colorbox[RGB]{252,169,141}{duck\vphantom{fg}}\hspace*{0pt}\colorbox[RGB]{252,204,183}{had\vphantom{fg}}\hspace*{0pt}\colorbox[RGB]{252,201,180}{better\vphantom{fg}}\hspace*{0pt}\colorbox[RGB]{254,225,211}{get\vphantom{fg}}\hspace*{0pt}\colorbox[RGB]{253,223,209}{away\vphantom{fg}}\hspace*{0pt}\colorbox[RGB]{254,233,223}{to\vphantom{fg}}\hspace*{0pt}\colorbox[RGB]{254,231,221}{safety\vphantom{fg}}\hspace*{0pt}\colorbox[RGB]{253,218,202}{fast!\vphantom{fg}}\hspace*{0pt}\colorbox[RGB]{254,234,224}{Answer:\vphantom{fg}}\hspace*{0pt}

\\

\bottomrule
\end{tabular}
}
%\end{center}
% \vspace{-.1in}
\end{table*}

\subsubsection{Attention by Gradient}
Consider an input $x = [t_1^{(1)}, t_2^{(1)}, ..., t_n^{(k)}]$ comprised of $k$ words and $n$ tokens, where $t_i^{(j)}$ represents the $i$-th token belonging to word $w_j$, and let $y$ be the corresponding label. Initially, LLM $f_\theta$ decomposes each word into tokens. Thus, tokens that correspond to the same word need to be concatenated, let the mapping function $w_j = M(t_i^{(j)})$. We first compute the gradient of each token according to:
\begin{align}
g_{t_i^{(j)}} = \frac{\partial \mathcal{L}[f_\theta(x), y]}{\partial t_i^{j}}.
\end{align}

Once we obtain the gradients, we compute the word-level gradient by summing the token-level gradients corresponding to each word:
\begin{align}
g_{w_j} = \sum \limits_{i \in {0, 1, ..., n}} g_{t_i^{(j)}} \
\text{s.t. } M(t_i^{(j)}) = w_j.
\end{align}

Finally, we calculate the $l_2$ norm of each word's gradient, followed by min-max normalization to produce a score $s_{w_j}$ for each word:
\begin{align}
s_{w_j} = \frac{|| g_{w_j} ||2 - \min g_{w_i}}{ \max g_{w_i} - \min g_{w_i} }.
\end{align}

\subsubsection{Attention by Deletion}

Attention by deletion is a prevalent method used in black-box textual attacks to determine the significance of each word in the input. Given an input $x$ with the $i$-th word $w_i$ deleted, denoted as $\hat{x}^{(i)}$, the importance score of $w_i$ can be computed by taking the absolute difference of the loss function $\mathcal{L}$ evaluated at the complete input $x$ and the altered input $\hat{x}^{(i)}$:
\begin{align}
s_{w_j} = | \mathcal{L}[f_\theta(x), y] - \mathcal{L}[f_\theta(\hat{x}^{(i)}). y] |
\end{align}

This raw score is then normalized using min-max normalization, yielding a final score $s_{w_j}$ for each word:
\begin{align}
s_{w_j} = \frac{s_{w_j} - \min s_{w_i}}{\max s_{w_i} - \min s_{w_i}}.
\end{align}

% \section{Word frequency analysis}
% \label{sec-append-word-frequency}

% \figurename~\ref{fig: mrpc word frequency} illustrates the distribution of word frequency as processed by the T5 model within the MRPC dataset. This distribution reveals a significant overlap in the occurrence of words in both robust and vulnerable prompts. This commonality suggests that the robustness or vulnerability of a prompt is not solely determined by the presence of specific words but potentially more related to their contextual use or placement within the prompt. It underscores the complexity and challenge of discerning between robust and vulnerable words based solely on frequency counts. This observation motivates the need for further investigation into additional factors, such as semantic coherence or syntactic structures, that may contribute to the robustness of prompts. For more granular analysis on word frequency, please refer to our detailed reports available in our Github repository\footnote{\url{https://github.com/microsoft/promptbench}}.

\subsection{Analysis on sentence-level attacks}
\label{stresstest analysis}

The phenomenon where sentence-level attacks occasionally improve the performance of \llms is an intriguing aspect of our study. Our attention analysis revealed distinct behaviors when models are subjected to StressTest and CheckList attacks. Specifically, when juxtaposed with other adversarial prompts, sentence-level attacks sometimes lead the model to hone in more acutely on pertinent keywords in the question and the labels. This is contrary to the expected behavior. As illustrated in \tablename~\ref{table: attention correctly classified adv prompts}, introducing an ostensibly unrelated sequence, such as `\prompt{and true is true}', heightens the \llms's focus on the `\prompt{not\_entailment}' label. Simultaneously, the model continues to attend to salient terms like `\prompt{minnow}' and `\prompt{duck}', ultimately culminating in a correct prediction.

\section{Details on Word frequency Analysis}
\label{sec-exp-word-frequency}

\begin{figure}[htbp]
    \centering
    \includegraphics[width=0.7\textwidth]{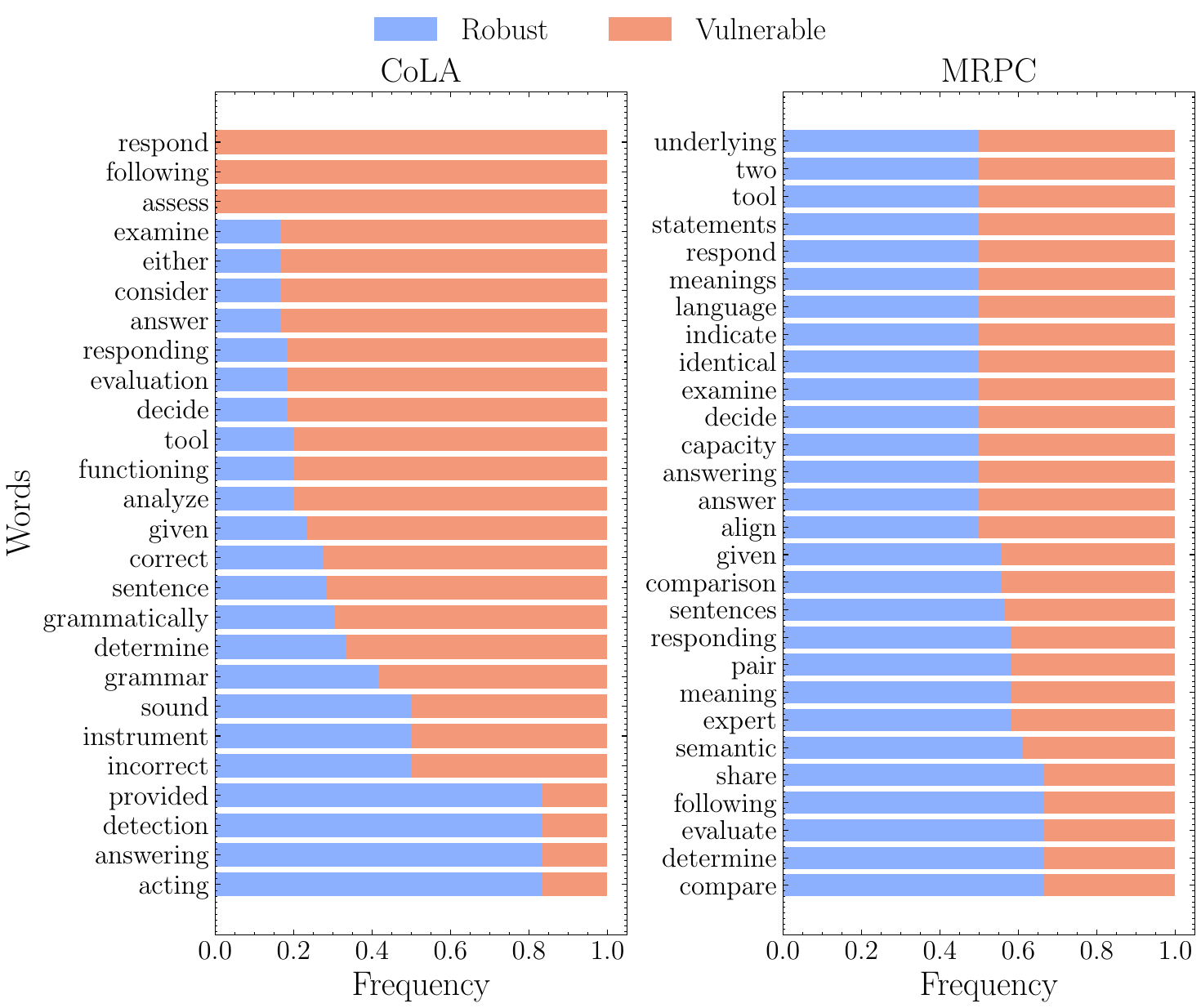}
    \caption{Frequency analysis for robust and vulnerable words on the CoLA (left) and MRPC (right) tasks.}
    \label{fig-word}
\end{figure}

Identifying the frequent patterns in prompts that may affect robustness is essential to both researchers and end-users.
We perform an initial analysis on word frequency.
We divide prompts into two categories: Vulnerable prompts, causing a performance drop of over $10\%$, and Robust prompts, with a performance drop of $10\%$ or less. Then, we collect all the words appeared in these prompts, and calculate the robust word frequency $f_{w_i}$ of word $w_i$ as $f^{robust}_{w_i} = \frac{n^{robust}_{w_i}}{n^{robust}_{w_i} + n^{vulnerable}_{w_i}},$ where $n^{robust}_{w_i}$ and $n^{vulnerable}_{w_i}$ denote the occurrences of $w_i$ in robust and vulnerable prompts, respectively. We primarily analyzed the adversarial prompts of CoLA and MRPC datasets generated by the T5-large model.
The word frequency results of these two datasets are presented in \figurename~\ref{fig-word}.

In our examination of adversarial robustness in large language models, we identified that word-specific resilience to attacks is not uniform. Specifically, within the COLA dataset, prompts incorporating terms such as ``\prompt{acting}'', ``\prompt{answering}'', and ``\prompt{detection}'' displayed greater resistance to adversarial perturbations. In contrast, those with words like ``\prompt{analyze}'', ``\prompt{answer}'', and ``\prompt{assess}'' were notably more susceptible. Yet, an analysis of the MRPC dataset demonstrated a significant overlap in the frequency of words present in both robust and vulnerable prompts. This overlap challenges the notion that specific words inherently determine a prompt's resilience to adversarial attacks.

Our findings underscore that the resilience of a prompt is intricately tied to the contextual use of words, rather than the mere presence of certain terms. This complexity suggests that factors beyond word frequency, such as semantic coherence and syntactic structures, might be instrumental in determining robustness. This knowledge is valuable as it can influence future research on the robustness of large language models, provide guidance for crafting more resistant prompts, and facilitate the creation of defensive mechanisms against adversarial prompt attacks. It's essential to emphasize that our observations are rooted in the current scope of models and datasets. Furthermore, the robustness or vulnerability of words remains deeply context-dependent. Hence, direct determination of word robustness without considering the broader context may lead to oversimplified or inaccurate conclusions.

\section{Defenses}
\label{sec-append-defense}

Now we discuss potential countermeasures for future research.
We categorize the robustness enhancement (i.e., defenses to adversarial prompts) approaches into three main axes: strategies in the training phase, input preprocessing, and downstream fine-tuning.

\subsection{Strategies in the training phase}
\textbf{Adversarial data integration.} Similar to adversarial training \citep{goodfellow2014explaining, kurakin2016adversarial}, integrating low-quality or intentionally perturbed data during the training and fine-tuning phases allows the model to acquaint itself with a broader range of inputs. This acclimatization aims to reduce the model's susceptibility to adversarial attacks, bolstering its resilience against malicious attempts that exploit such data nuances.

\textbf{Mixture of experts (MoE).} As discussed in Sec.~\ref{sec-transfer}, adversarial prompts exhibit transferability but constrained. Thus, one promising countermeasure is the utilization of diverse models~\citep{jacobs1991adaptive, shazeer2017outrageously, pang2019improving}, training them independently, and subsequently ensembling their outputs. The underlying premise is that an adversarial attack, which may be  effective against a singular model, is less likely to compromise the predictions of an ensemble comprising varied architectures. On the other hand, a prompt attack can also perturb a prompt based on an ensemble of \llms, which could enhance transferability.  

\subsection{Input preprocessing}
\textbf{Automated spelling verification.} Leveraging spelling checks \citep{kukich1992techniques, damerau1964technique} to maintain input fidelity can counteract basic adversarial techniques targeting typographical errors (character-level attacks) and inconsistencies (sentence level attacks).

\textbf{Semantic input rephrasing.} Semantic rephrasing \citep{fadaee2017data, ribeiro2018semantically} involves analyzing the meaning and intent behind a prompt. Using auxiliary models or rule-based systems to discern potentially adversarial or malicious intent in inputs could filter out harmful or misleading prompts.

\textbf{Historical context verification.} By maintaining a limited history of recent queries \citep{malhotra2015long, quadrana2017personalizing}, the system can identify patterns or sequences of inputs that might be part of a more extensive adversarial strategy. Recognizing and flagging suspicious input sequences can further insulate the LLM from coordinated adversarial attacks.

\subsection{Downstream fine-tuning}
% \wjd{Write something about downstream fine-tuning to defend; cite https://dl.acm.org/doi/abs/10.1145/3510003.3510191}

\textbf{Exploring fine-tuning techniques.} The fine-tuning phase is instrumental in refining the prowess of LLMs. Exploring more effective fine-tuning methodologies, which adjust based on the detected adversarial input patterns, can be pivotal. With the continuous evolution of adversarial threats, dynamic and adaptive fine-tuning remains a promising avenue. For example, only fine-tuning on relevant slicing technique~\citep{zhang2022remos}, model soup \citep{wortsman2022model}, fine-tuning then interpolating~\citep{wortsman2022robust}, etc.

\section{The visualization website for adversarial prompts}
\label{sec-append-vis}

In order to provide an interactive and user-friendly platform for visualizing and exploring adversarial prompts, we developed a web-based application using Streamlit hosted by Hugging Face and will be released in the future. 
% \url{https://huggingface.co/spaces/March07/PromptBench}.

\begin{figure}[htbp]
    \centering
    \includegraphics[width=.48\textwidth]{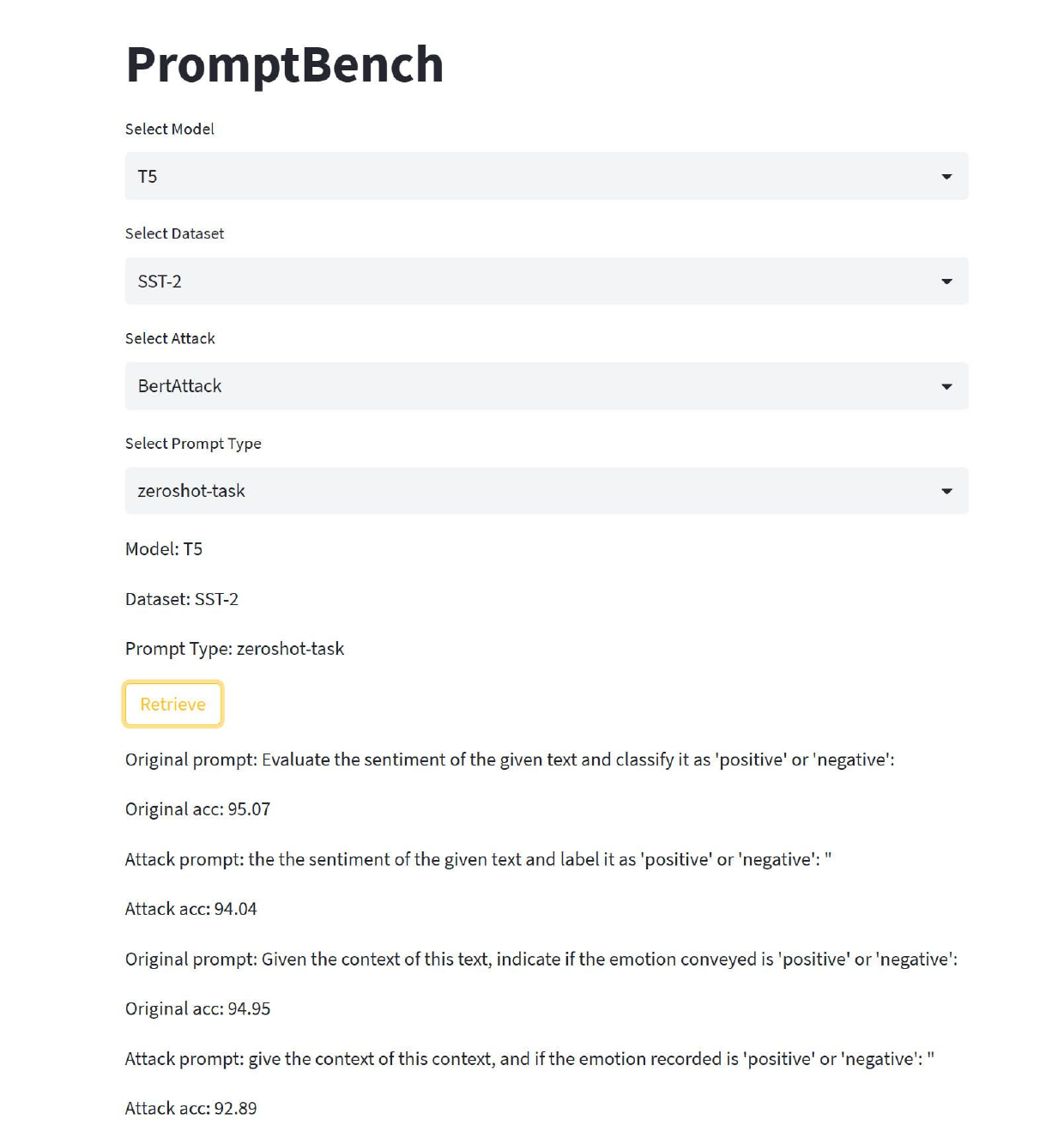}
    % \vspace{-.1in}
    \caption{The visualization website for adversarial prompts.}
    \label{fig: streamlit}
    % \vspace{-.1in}
\end{figure}

The visualization website, as shown in \figurename~\ref{fig: streamlit}, enables users to select from a variety of \llms (T5, Vicuna, UL2, ChatGPT), datasets (SST-2, CoLA, QQP, MRPC, MNLI, QNLI, RTE, WNLI, MMLU, SQuAD V2, IWSLT 2017, UN Multi, Math), prompt types (zeroshot-task, zeroshot-role, fewshot-task, and fewshot-role), and attacks (TextBugger, DeepWordBug, BertAttack, TextFooler, CheckList, StressTest, and Semantic). Based on the user's selection, the application generates adversarial prompts tailored to the chosen model, dataset, prompt type and attack.

\end{document}